%% file: 2018-iclr-cycle-adaptation.tex
\RequirePackage[log]{snapshot}
\documentclass{article} 
\usepackage{arxiv,times}
\usepackage{hyperref}
\usepackage{url}

\usepackage{epsfig} 
\usepackage{graphicx} 
\usepackage{amsmath} 
\usepackage{amssymb}

\usepackage[utf8]{inputenc} 
\usepackage[T1]{fontenc}    
\usepackage{booktabs}       
\usepackage{amsfonts}       
\usepackage{nicefrac}       
\usepackage{microtype}      

\usepackage{bbm}
\newcommand*\rot{\rotatebox{90}}
\usepackage{colortbl}
\definecolor{Gray}{gray}{0.85}
\newcolumntype{g}{>{\columncolor{Gray}} c}
\usepackage{caption}

\usepackage[colorinlistoftodos,prependcaption,textsize=tiny,disable]{todonotes}
\usepackage{xargs}

\newcommandx{\et}[2][1=]{\todo[linecolor=blue,backgroundcolor=blue!10,bordercolor=blue,#1]{Eric: #2}}
\newcommandx{\jh}[2][1=]{\todo[linecolor=green,backgroundcolor=green!10,bordercolor=green,#1]{Judy: #2}}
\newcommandx{\ks}[2][1=]{\todo[linecolor=red,backgroundcolor=red!10,bordercolor=red,#1]{Kate: #2}}
\newcommandx{\tp}[2][1=]{\todo[linecolor=orange,backgroundcolor=orange!10,bordercolor=orange,#1]{Taesung: #2}}
\newcommandx{\phil}[2][1=]{\todo[linecolor=blue,backgroundcolor=green!10,bordercolor=red,#1]{Phil: #2}}

\DeclareMathOperator*{\argmin}{arg\,min}

\begin{document}

\title{CyCADA: Cycle-Consistent Adversarial\\ Domain Adaptation}

%

\author{%
	Judy Hoffman, Eric Tzeng, Taesung Park, Jun-Yan Zhu\\
	BAIR, UC Berkeley\\
	\footnotesize{\texttt{\{jhoffman,etzeng,taesung\_park,junyanz\}@eecs.berkeley}}
	\And
	Phillip Isola\\
	OpenAI\thanks{Work done while at UC Berkeley}\\
	\small{\texttt{isola@eecs.berkeley}}
	\And
	Kate Saenko\\
	CS, Boston University\\
	\small{\texttt{saenko@bu}}\\ 
	\And
	Alexei A. Efros, Trevor Darrell\\
	BAIR, UC Berkeley\\
	\small{\texttt{\{efros,trevor\}@eecs.berkeley}}
}

\maketitle

\input{variables}
\input{abstract}
\input{introduction}

\input{related}

\input{method}
\input{experiments}

\input{conclusion}
\input{references}
\input{appendix}

\end{document}

%% file: variables.tex
\newcommand{\loss}{\mathcal{L}}
\newcommand{\lossgan}{\loss_\text{GAN}}
\newcommand{\losscyclegan}{\loss_\text{CycleGAN}}
\newcommand{\losscycleadapt}{\loss_\text{CycleAdapt}}

\newcommand{\StoT}{G_{S \rightarrow T}}
\newcommand{\TtoS}{G_{T \rightarrow S}}

\newcommand{\etal}{et al.}

%% file: abstract.tex
\begin{abstract}
Domain adaptation is critical for success in new, unseen environments.
Adversarial adaptation models applied in feature spaces discover domain invariant representations, but are difficult to visualize and sometimes fail to capture pixel-level and low-level domain shifts.
Recent work has shown that generative adversarial networks combined with cycle-consistency constraints are surprisingly effective at  mapping images between domains, even without the use of aligned image pairs.
We propose a novel discriminatively-trained Cycle-Consistent Adversarial Domain Adaptation model.
CyCADA adapts representations at both the pixel-level and feature-level, enforces cycle-consistency while leveraging a task loss, and does not require aligned pairs.  Our model can be applied in a variety of visual recognition and prediction settings.
We show new state-of-the-art results across multiple adaptation tasks, including digit classification and semantic segmentation of road scenes demonstrating transfer from synthetic to real world domains.

\end{abstract}

%% file: introduction.tex
\section{Introduction}

Deep neural networks excel at learning from large amounts of data, but can be poor at generalizing  learned knowledge to new datasets or environments.
Even a slight departure from a network's training domain can cause it to make spurious predictions and significantly hurt its performance~\citep{tzeng_cvpr17}.
The visual domain shift from non-photorealistic synthetic data to real images presents an even more significant challenge.
While we would like to train models on large amounts of synthetic data such as data collected from graphics game engines, such models fail to generalize to real-world imagery.
For example, a state-of-the-art semantic segmentation model trained on synthetic dashcam data fails to segment the road in real images, and its overall per-pixel label accuracy drops from 93\% (if trained on real imagery) to 54\% (if trained only on synthetic data, see Table~\ref{fig:gta-cityscapes}).

Feature-level unsupervised domain adaptation methods address this problem by aligning the features extracted from the network across the source (e.g. synthetic) and target (e.g. real) domains, without any labeled target samples.
Alignment typically involves minimizing some measure of distance between the source and target feature distributions, such as maximum mean discrepancy~\citep{long_icml15}, correlation distance~\citep{sun_taskcv16}, or adversarial discriminator accuracy ~\citep{ganin_icml15,tzeng_cvpr17}.
This class of techniques suffers from two main limitations. First, aligning marginal distributions does not enforce any semantic consistency, e.g. target features of a car may be mapped to source features of a bicycle. Second, alignment at higher levels of a deep representation can fail to model aspects of low-level appearance variance which are crucial for the end visual task. 

\begin{figure}[t]	
	\centering
	\includegraphics[width=\linewidth]{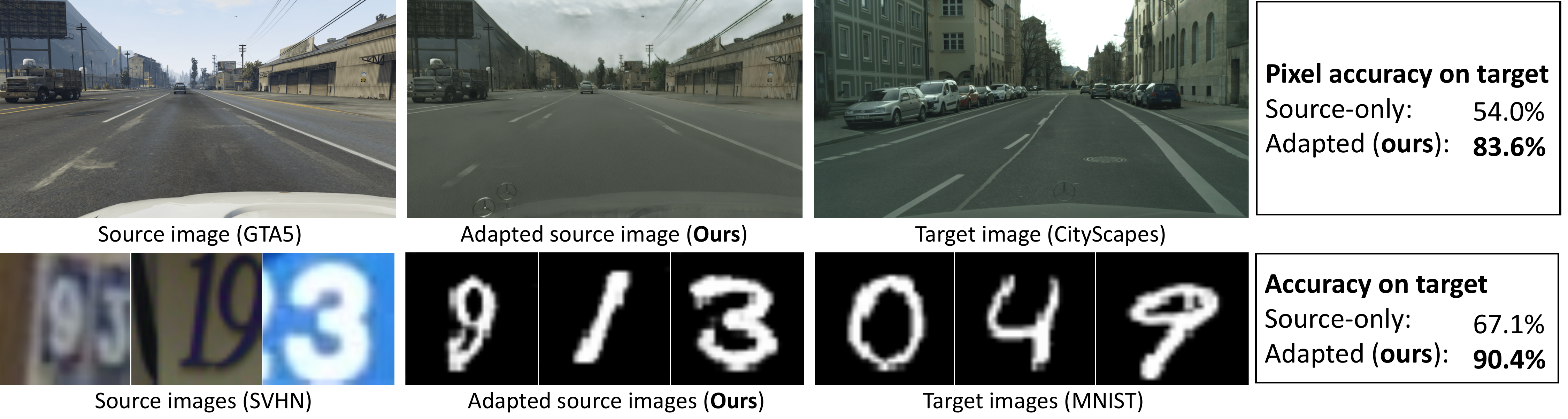}
	\caption{We propose CyCADA, an adversarial unsupervised adaptation algorithm which uses cycle and semantic consistency to perform adaptation at multiple levels in a deep network. Our model provides significant performance improvements over source model baselines.}
	\label{fig:teaser}
\end{figure}

Generative pixel-level domain adaptation models perform similar distribution alignment---not in feature space but rather in raw pixel space---translating source data to the ``style'' of a target domain. Recent methods can learn to translate images given only unsupervised data from both domains~\citep{bousmalis_cvpr17,liu_arxiv16,shrivastava_cvpr17}.
The results are visually compelling, but such image-space models have only been shown to work for small image sizes and limited domain shifts. A more recent approach \citep{bousmalis_arxiv17_robotic} was applied to larger (but still not high resolution) images, but in a visually controlled image for robotic applications. 
Furthermore, they also do not necessarily preserve content: while the translated image may ``look'' like it came from the right domain, crucial semantic information may be lost. For example, a model adapting from line-drawings to photos could learn to make a line-drawing of a cat look like a photo of a dog.

How can we encourage the model to preserve semantic information in the process of distribution alignment? In this paper, we explore a simple yet powerful idea: give an additional objective to the model to reconstruct the original data from the adapted version. Cycle-consistency was recently proposed in a cross-domain image generation GAN model, CycleGAN~\citep{zhu_arxiv17}, which showed transformative image-to-image generation results, but was agnostic to any particular task.

We propose Cycle-Consistent Adversarial Domain Adaptation (CyCADA), which adapts representations at both the pixel-level and feature-level while enforcing local and global structural consistency through pixel cycle-consistency and semantic losses.
CyCADA unifies prior feature-level \citep{ganin_icml15,tzeng_cvpr17} and image-level \citep{liu_arxiv16,bousmalis_cvpr17,shrivastava_cvpr17} adversarial domain adaptation methods together with cycle-consistent image-to-image translation techniques \citep{zhu_arxiv17}, as illustrated in Table~\ref{table:compare-methods}.
It is applicable across a range of deep architectures and/or representation levels, and 
has several advantages over existing unsupervised domain adaptation methods. We use a reconstruction (cycle-consistency) loss to encourage the cross-domain transformation to preserve local structural information and a semantic loss to enforce semantic consistency.

We apply our CyCADA model to the task of digit recognition across domains and the task of semantic segmentation of urban scenes across domains. Experiments show that our model achieves state of the art results on digit adaptation, cross-season adaptation in synthetic data, and on the challenging synthetic-to-real scenario. In the latter case, it improves per-pixel accuracy from 54\% to 82\%, nearly closing the gap to the target-trained model.

Our experiments confirm  that domain adaptation can benefit greatly from cycle-consistent pixel transformations, and that this is especially important for pixel-level semantic segmentation with contemporary FCN architectures. Further, we show that adaptation at both the pixel and representation level can offer complementary improvements with joint pixel-space and feature adaptation leading to the highest performing model for digit classification tasks. 

\begin{table}[t]
\resizebox{\textwidth}{!}{
\begin{tabular}{lcccc}
	\toprule
    & Pixel  & Feature  & Semantic  & Cycle \\
    & Loss & Loss & Loss & Consistent\\
    \midrule
    CycleGAN~\citep{zhu_arxiv17} & \checkmark & & & \checkmark\\
    Feature Adapt ~\citep{ganin_icml15,tzeng_cvpr17}& & \checkmark & \checkmark\\
    Pixel Adapt~\citep{dtn,bousmalis_cvpr17}& \checkmark & &\checkmark &  \\
    CyCADA & \checkmark & \checkmark & \checkmark & \checkmark\\
    \bottomrule
\end{tabular}
}
\caption{Our model, CyCADA, may use pixel, feature, and semantic information during adaptation while learning an invertible mapping through cycle consistency. }
\label{table:compare-methods}
\end{table}

%% file: related.tex
\section{Related Work}
The problem of visual domain adaptation was introduced along with a pairwise metric transform solution by \citet{saenko_eccv10} and was further popularized by the broad study of visual dataset bias~\citep{efros_cvpr11}.
Early deep adaptive works focused on feature space alignment through minimizing the distance between first or second order feature space statistics of the source and target~\citep{tzeng_arxiv15,long_icml15}. These latent distribution alignment approaches were further improved through the use of domain adversarial objectives whereby a domain classifier is trained to distinguish between the source and target representations while the domain representation is learned so as to maximize the error of the domain classifier. The representation is optimized using the standard minimax objective~\citep{ganin_icml15}, the symmetric confusion objective~\citep{tzeng_iccv15}, or the inverted label objective~\citep{tzeng_cvpr17}.
Each of these objectives is related to the literature on generative adversarial networks~\citep{goodfellow_nips14} and follow-up work for improved training procedures for these networks~\citep{salimans_arxiv16, arjovsky_arxiv17}.

The feature-space adaptation methods described above focus on modifications to the discriminative representation space. In contrast, other recent methods have sought adaptation in the pixel-space using various generative approaches. One advantage of pixel-space adaptation, as we have shown, is that the result may be more human interpretable, since an image from one domain can now be visualized in a new domain.
CoGANs~\citep{liu_arxiv16} jointly learn a source and target representation through explicit weight sharing of certain layers while each source and target has a unique generative adversarial objective. \citet{ghifary_eccv16} uses an additional reconstruction objective in the target domain to encourage alignment in the unsupervised adaptation setting.

In contrast, another approach is to directly convert the target image into a source style image (or visa versa), largely based on  Generative Adversarial Networks (GANs)~\citep{goodfellow_nips14}.
Researchers have successfully applied GANs to various applications such as image generation~\citep{denton2015deep,radford2015unsupervised,zhao2016energy}, image editing~\citep{zhu2016generative} and feature learning~\citep{salimans2016improved,donahue2016adversarial}.   Recent work \citep{isola2016image, sangkloy2016scribbler,karacan2016learning} adopt conditional GANs~\citep{mirza_arxiv14} for these image-to-image translation problems~\citep{isola2016image}, but they require input-output image pairs for training, which is in general not available in domain adaptation problems.

There also exist lines of work where such training pairs are not given.
\citet{yoo_eccv16} learns a source to target encoder-decoder along with a generative adversarial objective on the reconstruction which is is applied for predicting the clothing people are wearing.
The Domain Transfer Network~\citep{taigman_iclr17} trains a generator to transform a source image into a target image by enforcing consistency in the embedding space.
\citet{shrivastava_cvpr17} instead uses an L1 reconstruction loss to force the generated target images to be similar to their original source images.%
This works well for limited domain shifts where the domains are similar in pixel-space, but can be too limiting for settings with larger domain shifts.
\citet{bousmalis_cvpr17} use a content similarity loss to ensure the generated target image is similar to the original source image; however, this requires prior knowledge about which parts of the image stay the same across domains (e.g. foreground).
Our method does not require pre-defining what content is shared between domains and instead simply translates images back to their original domains while ensuring that they remain identical to their original versions.
BiGAN~\citep{donahue2016adversarial} and ALI~\citep{dumoulin2016adversarially} take an approach of simultaneously learning the transformations between the pixel and the latent space.
More recently, Cycle-consistent Adversarial Networks (CycleGAN)~\citep{zhu_arxiv17} produced compelling image translation results such as generating photorealistic images from impressionism paintings or transforming horses into zebras at high resolution using the cycle-consistency loss.
This loss was simultaneously proposed by ~\cite{yi2017dualgan} and ~\cite{kim_arxiv17} to great effect as well.
Our motivation comes from such findings about the effectiveness of the cycle-consistency loss. 

Few works have explicitly studied visual domain adaptation for the semantic segmentation task. Adaptation across weather conditions in simple road scenes was first studied by \citet{levinkov_iccv13}. More recently, a convolutional domain adversarial based approached was proposed for more general drive cam scenes and for adaptation from simulated to real environments~\citep{hoffman_arxiv16}.
\citet{ros_arxiv16} learns a multi-source model through concatenating all available labeled data and learning a single large model and then transfers to a sparsely labeled target domain through distillation~\citep{hinton_arxiv15}.
\citet{chen_iccv17} use an adversarial objective to align both global and class-specific statistics, while mining additional temporal data from street view datasets to learn a static object prior.
\citet{zhang_iccv17} instead perform segmentation adaptation by aligning label distributions both globally and across superpixels in an image.

%% file: method.tex
\section{Cycle-Consistent Adversarial Domain Adaption}

\begin{figure}
  \includegraphics[width=\textwidth]{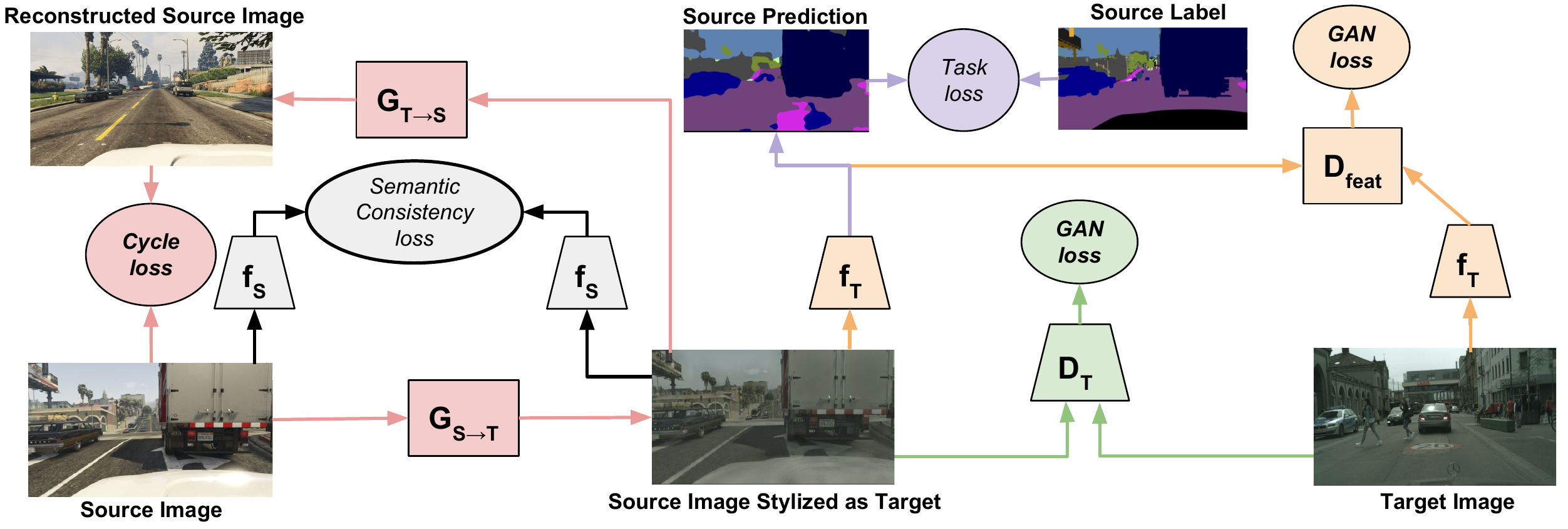}
  \caption{
    Cycle-consistent adversarial adaptation of pixel-space inputs.
    By directly remapping source training data into the target domain, we remove the low-level differences between the domains, ensuring that our task model is well-conditioned on target data. We depict here the image-level GAN loss (\textcolor{green}{green}), the feature level GAN loss (\textcolor{orange}{orange}), the source and target semantic consistency losses (black), the source cycle loss (\textcolor{red}{red}), and the source task loss (\textcolor{purple}{purple}). For clarity the target cycle is omitted. 
  }
  \label{fig:implementation}
\end{figure}

We consider the problem of unsupervised adaptation, where we are provided source data $X_S$, source labels $Y_S$, and target data $X_T$, but no target labels.
The goal is to learn a model $f$ that can correctly predict the label for the target data $X_T$.

We can begin by simply learning a source model $f_S$ that can perform the task on the source data.
For $K$-way classification with a cross-entropy loss, this corresponds to
\begin{align}
  \loss_{\text{task}}(f_S, X_S, Y_S) = 
  		- \mathbb{E}_{\small{(x_s, y_s) \sim (X_S, Y_S)} }
        \sum_{k=1}^K \mathbbm{1}_{[k = y_s]} \log \left( \sigma(f_S^{(k)}(x_s)) \right) 
\end{align}
where $\sigma$ denotes the softmax function. 
However, while the learned model $f_S$ will perform well on the source data, typically domain shift between the source and target domain leads to reduced performance when evaluating on target data.
To mitigate the effects of domain shift, we follow previous adversarial adaptation approaches and learn to map samples across domains such that an adversarial discriminator is unable to distinguish the domains.
By mapping samples into a common space, we enable our model to learn on source data while still generalizing to target data.

To this end, we introduce a mapping from source to target $\StoT$ and train it to produce target samples that fool an adversarial discriminator $D_T$.
Conversely, the adversarial discriminator attempts to classify the real target data from the source target data.
This corresponds to the loss function
\begin{align}
  \label{eq:gan-loss}
  \lossgan( G_{S \rightarrow T}, D_T, X_T, X_S) = \mathbb{E}_{x_t \sim X_T}\left[\log D_T(x_t)\right]
    + \mathbb{E}_{x_s \sim X_S}\left[\log(1 - D_T(\StoT(x_s)))\right] 
\end{align}
This objective ensures that $\StoT$, given source samples, produces convincing target samples.
In turn, this ability to directly map samples between domains allows us to learn a target model $f_T$ by minimizing $\loss_{\text{task}}(f_T, \StoT(X_S), Y_S)$ (see Figure~\ref{fig:implementation} green portion).

However, while previous approaches that optimized similar objectives have shown effective results, in practice they can often be unstable and prone to failure.
Although the GAN loss in Equation~\ref{eq:gan-loss} ensures that $\StoT(x_s)$ for some $x_s$ will resemble data drawn from $X_T$, there is no way to guarantee that $\StoT(x_s)$ preserves the structure  or content of the original sample $x_s$.

In order to encourage the source content to be preserved during the conversion process, we impose a cycle-consistency constraint on our adaptation method~\citep{zhu_arxiv17,yi2017dualgan,kim_arxiv17} (see Figure~\ref{fig:implementation} red portion).
To this end, we introduce another mapping from target to source $\TtoS$ and train it according to the same GAN loss $\lossgan(\TtoS, D_S, X_S, X_T)$.
We then require that mapping a source sample from source to target and back to the source reproduces the original sample, thereby enforcing cycle-consistency.
In other words, we want $\TtoS(\StoT(x_s)) \approx x_s$ and $\StoT(\TtoS(x_t)) \approx x_t$.
This is done by imposing an L1 penalty on the reconstruction error, which is referred to as the \emph{cycle-consistency loss}:
\begin{align}
  \label{eq:cycle-loss}
  \mathcal{L}_\text{cyc}(\StoT,  \TtoS, X_S, X_T) &=
   \mathbb{E}_{x_s \sim X_S}\left[||\TtoS(\StoT(x_s)) - x_s||_1\right] \\
  &+  \mathbb{E}_{x_t \sim X_T}\left[||\StoT(\TtoS(x_t)) - x_t||_1\right].\nonumber
\end{align}
%
Additionally, as we have access to source labeled data, we explicitly encourage high semantic consistency before and after image translation. 
We pretrain a source task model $f_S$, fixing the weights, we use this model as a noisy labeler by which we encourage an image to be classified in the same way after translation as it was before translation according to this classifier. Let us define the predicted label from a fixed classifier, $f$, for a given input $X$ as $p(f,X) = \text{arg} \max(f(X))$. Then we can define the semantic consistency before and after image translation as follows:
\begin{align}
	\label{eq:semantic}
	\mathcal{L}_\text{sem}(\StoT, \TtoS, X_S, X_T, f_S) &=
	 \mathcal{L}_{\text{task}}(f_S, \TtoS(X_T), p(f_S, X_T))  \\
	&+  \mathcal{L}_{\text{task}}(f_S, \StoT(X_S), p(f_S, X_S)) \nonumber 
\end{align}
See Figure~\ref{fig:implementation} black portion. This can be viewed as analogously to content losses in style transfer ~\citep{gatys2016image} or in pixel adaptation \citep{dtn}, where the shared content to preserve is dictated by the source task model $f_S$.

We have thus far described an adaptation method which combines cycle consistency, semantic consistency, and adversarial objectives to produce a final target model. As a pixel-level method, the adversarial objective consists of a discriminator which distinguishes between two image sets, e.g. transformed source and real target image. Note that we could also consider a feature-level method which discriminates between the features or semantics from two image sets as viewed under a task network. This would amount to an additional feature level GAN loss (see Figure~\ref{fig:implementation} orange portion):
\begin{equation}
	\lossgan(f_T, D_\text{feat}, f_S(\StoT(X_S)), X_T).\label{eq:adversarial-feature}
\end{equation}

Taken together, these loss functions form our complete objective:
\begin{align}
  \label{eq:cycada}
  \loss_{\text{CyCADA}}&(f_T, X_S, X_T, Y_S, \StoT, \TtoS, D_S, D_T) \\
  &= \loss_{\text{task}}(f_T, \StoT(X_S), Y_S) \nonumber \\
  &+ \lossgan(\StoT, D_T, X_T, X_S)
  + \lossgan(\TtoS, D_S, X_S, X_T) \nonumber \\
  &+ \lossgan(f_T, D_\text{feat}, f_S(\StoT(X_S)), X_T)\nonumber \\
  &+ \loss_\text{cyc}(\StoT, \TtoS, X_S, X_T)
  + \mathcal{L}_\text{sem}(\StoT, \TtoS, X_S, X_T, f_S). \nonumber
\end{align}

This ultimately corresponds to solving for a target model $f_T$ according to the optimization problem
\begin{equation}
   f_T^* = \argmin_{f_T} \min_{\substack{\StoT\\ \TtoS}} \max_{D_S, D_T} \loss_\text{CyCADA}(\small{f_T, X_S, X_T, Y_S, \StoT, \TtoS, D_S, D_T}).
\end{equation}

We have introduced a method for unsupervised adaptation which generalizes adversarial objectives to be viewed as operating at the pixel or feature level. In addition, we introduce the use of cycle-consistency together with semantic transformation constraints to guide the mapping from one domain to another. 
In this work, we apply CyCADA to both digit adaptation and to semantic segmentation. We implement $G$ as a pixel-to-pixel convnet, $f$ as a convnet classifier or a Fully-Convolutional Net (FCN) and $D$ as a convnet with binary outputs.

%% file: experiments.tex
\section{Experiments}
\label{sec:experiments}

We evaluate CyCADA on several unsupervised adaptation scenarios. We first focus on adaptation for digit classification using the MNIST~\citep{lecun_ieee98}, USPS, and Street View House Numbers (SVHN)~\citep{netzer_nips11} datasets. After which we present results for the task of semantic image segmentation, using the SYNTHIA~\citep{ros_cvpr16}, GTA~\citep{richter_eccv16} and CityScapes~\citep{cordts_cvpr16} datasets. 

\subsection{Digit Adaptation}
We evaluate our method across the adaptation shifts of USPS to MNIST, MNIST to USPS, and SVHN to MNIST, using the full training sets during learning phases and evaluating on the standard test sets. We report classification accuracy for each shift in Table~\ref{table:digits} and find that our method outperforms competing approaches on average. 
The classifier for our method for all digit shifts uses a variant of the LeNet architecture (see \ref{sec:digit-details} for full implementation details). 
Note that the recent pixel-da method by \citet{bousmalis_cvpr17} presents results for only the MNIST to USPS shift and reports 95.9\% accuracy, while our method achieves 95.6\% accuracy. However, the pixel-da approach cross validates with some labeled data which is not an equivalent evaluation setting. 

\textbf{Ablation: Pixel vs Feature Level Transfer.} We begin by evaluating the contribution of the pixel space and feature space transfer. 
We find that in the case of the small domain shifts between USPS and MNIST, the pixel space adaptation by which we train a classifier using images translated using CycleGAN~\citep{zhu_arxiv17}, performs very well, outperforming or comparable to prior adaptation approaches. 
Feature level adaptation offers a small benefit in this case of a small pixel shift. 
However, for the more difficult shift of SVHN to MNIST, we find that feature level adaptation outperforms the pixel level adaptation, and importantly, both may be combined to produce an overall model which outperforms all competing methods. 

\input{tables/digits.tex}
\begin{figure}
	\centering
	\begin{tabular}{cc}
	\includegraphics[height=.1\linewidth]{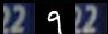} &
	\includegraphics[height=.1\linewidth]{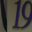}
	\includegraphics[height=.1\linewidth]{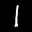}
	\includegraphics[height=.1\linewidth]{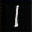}
	\\
	\includegraphics[height=.1\linewidth]{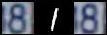} &
	\includegraphics[height=.1\linewidth]{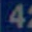}
	\includegraphics[height=.1\linewidth]{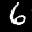}
	\includegraphics[height=.1\linewidth]{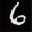}\\
	(a) Without Semantic Loss & (b) Without Cycle Loss
	\end{tabular}
	\caption{\textbf{Ablation: Effect of Semantic or Cycle Consistency} Examples of translation failures without the semantic consistency loss. Each triple contains the original SVHN image (\textit{left}), the image translated into MNIST style (\textit{middle}), and the image reconstructed back into SVHN (\textit{right}). (a) Without semantic loss, both the GAN and cycle constraints are satisfied (translated image matches MNIST style and reconstructed image matches original), but the image translated to the target domain lacks the proper semantics. (b) Without cycle loss, the reconstruction is not satisfied and though the semantic consistency leads to some successful semantic translations (\textit{top}) there are still cases of label flipping (\textit{bottom}).}
	\label{fig:cyclegan_fail}
\end{figure}

\textbf{Ablation: No Semantic Consistency.} We experiment without the addition of our semantic consistency loss and find that the standard unsupervised CycleGAN approach diverged when training SVHN to MNIST often suffering from random label flipping.
 Figure~\ref{fig:cyclegan_fail}(a) demonstrates two examples where cycle constraints alone fail to have the desired behavior for our end task. An SVHN image is mapped to a convincing MNIST type image and back to a SVHN image with correct semantics. However, the MNIST-like image has mismatched semantics. Our modified version, which uses the source labels to train a weak classification model which can be used to enforce semantic consistency before and after translation, resolves this issue and produces strong performance.
 
 \textbf{Ablation: No Cycle Consistency.} We study the result when learning without the cycle consistency loss. First note that there is no reconstruction guarantee in this case, thus in Figure~\ref{fig:cyclegan_fail}(b) we see that the translation back to SVHN fails. In addition, we find that while the semantic loss does encourage correct semantics it relies on the weak source labeler and thus label flipping still occurs (see right image triple).

\subsection{Semantic Segmentation Adaptation}
The task is to assign a semantic label to each pixel in the input image, e.g. $road$, $building$, etc.
We limit our evaluation to the unsupervised adaptation setting, where labels are only available in the source domain, but we are evaluated solely on our performance in the target domain.

For each experiment, we use three metrics to evaluate performance.
Let $n_{ij}$ be the number of pixels of class $i$ predicted as class $j$, let $t_i = \sum_j n_{ij}$ be the total number of pixels of class $i$, and let $N$ be the number of classes.
Our three evaluation metrics are, mean intersection-over-union (mIoU), frequency weighted intersection-over-union (fwIoU), and pixel accuracy, which are defined as follows: \\
mIoU $=\frac{1}{N} \cdot \frac{\sum_i n_{ii}}{t_i + \sum_j n_{ji} - n_{ii}}$, fwIoU $=\frac{1}{\sum_k t_k} \cdot \frac{\sum_i n_{ii}}{t_i + \sum_j n_{ji} - n_{ii}}$, pixel acc. $=\frac{\sum_i n_{ii}}{\sum_i t_i}$.

Cycle-consistent adversarial adaptation is general and can be applied at any layer of a network. 
Since optimizing the full CyCADA objective in Equation~\ref{eq:cycada} end-to-end is memory-intensive in practice, we train our model in stages.
First, we perform image-space adaptation and map our source data into the target domain.
Next, using the adapted source data with the original source labels, we learn a task model that is suited to operating on target data.
Finally, we perform another round of adaptation between the adapted source data and the target data in feature-space, using one of the intermediate layers of the task model. Additionally, we do not use the semantic loss for the segmentation experiments as it would require loading generators, discriminators, and an additional semantic segmenter into memory all at once for two images. We did not have the required memory for this at the time of submission, but leave it to future work to deploy model parallelism or experiment with larger GPU memory.

For our first evaluation, 
we consider the SYNTHIA dataset~\citep{ros_cvpr16}, which contains synthetic renderings of urban scenes.
We use the SYNTHIA video sequences, which are rendered across a variety of environments, weather conditions, and lighting conditions.
This provides a synthetic testbed for evaluating adaptation techniques.
For comparison with previous work, in this work we focus on adaptation between seasons.
We use only the front-facing views in the sequences so as to mimic dashcam imagery, and adapt from fall to winter.
The subset of the dataset we use contains 13 classes and consists of 10,852 fall images and 7,654 winter images.

To further demonstrate our method's applicability to real-world adaptation scenarios, we also evaluate our model in a challenging synthetic-to-real adaptation setting.
For our synthetic source domain, we use the GTA5 dataset~\citep{richter_eccv16} extracted from the game Grand Theft Auto V, which contains 24966 images. 
We consider adaptation from GTA5 to the real-world Cityscapes dataset~\citep{cordts_cvpr16}, from which we used 19998 images without annotation for training and 500 images for validation. 
Both of these datasets are evaluated on the same set of 19 classes, allowing for straightforward adaptation between the two domains.

Image-space adaptation also affords us the ability to visually inspect the results of the adaptation method.
This is a distinct advantage over opaque feature-space adaptation methods, especially in truly unsupervised settings---without labels, there is no way to empirically evaluate the adapted model, and thus no way to verify that adaptation is improving task performance.
Visually confirming that the conversions between source and target images are reasonable, while not a \emph{guarantee} of improved task performance, can serve as a sanity check to ensure that adaptation is not completely diverging. This process is diagrammed in Figure~\ref{fig:implementation}. For implementation details please see Appendix \ref{sec:ss-details}.

\newcommand{\myw}{3.5cm}
\newcommand{\myh}{1.8cm}
\subsubsection{Cross-season adaptation}

We start by exploring the abilities of pixel space adaptation alone (using FCN8s architecture) for the setting of adapting across seasons in synthetic data. For this we use the SYNTHIA dataset and adapt from fall to winter weather conditions. 
Typically in unsupervised adaptation settings it is difficult to interpret what causes the performance improvement after adaptation.
Therefore, we use this setting as an example where we may directly visualize the shift from fall to winter and inspect the intermediate pixel level adaptation result from our algorithm.
In Figure~\ref{fig:synthia} we show the result of pixel only adaptation as we generate a winter domain image (b) from a fall domain image (a), and visa versa (c-d). We may clearly see the changes of adding or removing snow. This visually interpretable result matches our expectation of the true shift between these domains and indeed results in favorable final semantic segmentation performance from fall to winter as shown in Table~\ref{table:synthia}. We find that CyCADA achieves state-of-the-art performance on this task with image space adaptation alone, however does not recover full supervised learning performance (train on target). Some example errors includes adding snow to the sidewalks, but not to the road, while in the true winter domain snow appears in both locations. However, even this mistake is interesting as it implies that the model is learning to distinguish road from sidewalk during pixel adaptation, despite the lack of pixel annotations.

\input{figs/synthia.tex}
\input{tables/synthia}
Cycle-consistent adversarial adaptation achieves state-of-the-art adaptation performance.
We see that under the fwIoU and pixel accuracy metrics, CyCADA approaches oracle performance, falling short by only a few points, despite being entirely unsupervised.
This indicates that CyCADA is extremely effective at correcting the most common classes in the dataset.
This conclusion is supported by inspection of the individual classes in Table~\ref{table:synthia}, where we see the largest improvement on common classes such as \emph{road} and \emph{sidewalk}.

\input{figs/gta-cityscapes-seg.tex}

\input{tables/gta-cityscapes.tex}
\input{figs/gta-cityscapes.tex}

\subsubsection{Synthetic to real adaptation}

To evaluate our method's applicability to real-world adaptation settings, we investigate adaptation from synthetic to real-world imagery.
The results of this evaluation are presented in Table~\ref{table:gta-cityscapes} with qualitative results shown in Figure~\ref{fig:gta-cityscapes-seg}.
Once again, CyCADA achieves state-of-the-art results, recovering approximately 40\% of the performance lost to domain shift.
CyCADA also improves or maintains performance on all 19 classes.
Examination of fwIoU and pixel accuracy as well as individual class IoUs reveals that our method performs well on most of the common classes.
Although some classes such as \emph{train} and \emph{bicycle} see little or no improvement, we note that those classes are poorly represented in the GTA5 data, making recognition very difficult. We compare our model against \citet{shrivastava_cvpr17} for this setting, but found this approach did not converge and resulted in worse performance than the source only model (see Appendix for full details).

We visualize the results of image-space adaptation between GTA5 and Cityscapes in Figure~\ref{fig:gta-cityscapes}.
The most obvious difference between the original images and the adapted images is the saturation levels---the GTA5 imagery is much more vivid than the Cityscapes imagery, so adaptation adjusts the colors to compensate.
We also observe texture changes, which are perhaps most apparent in the road: in-game, the roads appear rough with many blemishes, but Cityscapes roads tend to be fairly uniform in appearance, so in converting from GTA5 to Cityscapes, our model removes most of the texture.
Somewhat amusingly, our model has a tendency to add a hood ornament to the bottom of the image, which, while likely irrelevant to the segmentation task, serves as a further indication that image-space adaptation is producing reasonable results.

%% file: tables/digits.tex
\begin{table}
  \centering
  \scriptsize
  \setlength{\tabcolsep}{1.35pt}
  \begin{tabular}{lccc}
    \toprule
	Model & 		MNIST $\rightarrow$ USPS & USPS $\rightarrow$ MNIST & SVHN $\rightarrow$ MNIST \\
	\midrule \midrule
	Source only & 	82.2 $\pm$ 0.8 &  69.6 $\pm$ 3.8 &	67.1 $\pm$ 0.6\\
	DANN~\citep{dann}&	-	&	-	&73.6\\
	DTN~\citep{dtn} &	-&		-&	84.4\\
	CoGAN~\citep{cogan} &	91.2 &	89.1 &	-\\
	ADDA~\citep{tzeng_cvpr17} & 	89.4 $\pm$ 0.2 &	90.1 $\pm$ 0.8 &	76.0 $\pm$ 1.8\\
	
        CyCADA pixel only & 	 \bf 95.6 $\pm$ 0.2 &	96.4 $\pm$ 0.1&	70.3 $\pm$ 0.2\\
	CyCADA pixel+feat & 	\bf  95.6 $\pm$ 0.2 & 	\bf	96.5 $\pm$ 0.1 &	\bf	90.4 $\pm$ 0.4\\
	\midrule
	Target only & 	96.3 $\pm$ 0.1 & 	99.2 $\pm$ 0.1&	99.2 $\pm 0.1$\\
    \bottomrule
  \end{tabular}
  \caption{
    \textbf{Unsupervised domain adaptation across digit datasets.} Our model is competitive with or outperforms state-of-the-art models for each shift. For the difficult shift of SVHN to MNIST we also note that feature space adaptation provides additional benefit beyond the pixel-only adaptation. 
    }
  \label{table:digits}
\end{table}

%% file: figs/synthia.tex
\begin{figure*}
  \centering
  \setlength{\tabcolsep}{2.0pt}
  \begin{tabular}{cccc}
    \includegraphics[width=0.24\textwidth]{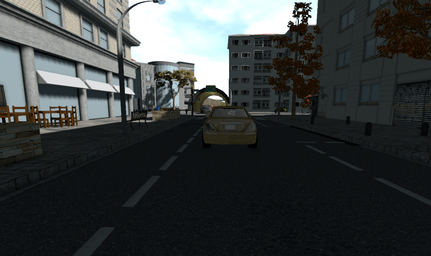} &
    \includegraphics[width=0.24\textwidth]{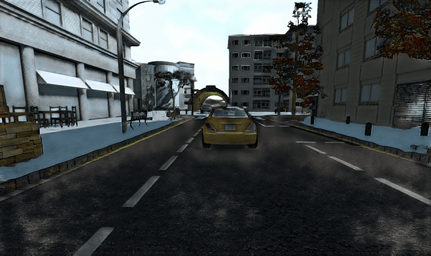} &
    \includegraphics[width=0.24\textwidth]{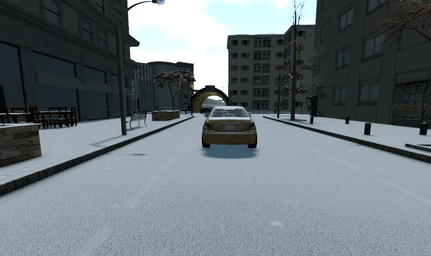} &
    \includegraphics[width=0.24\textwidth]{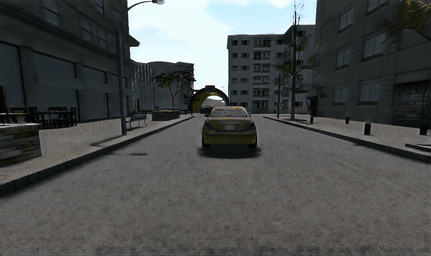}
    \\
	   (a) Fall & (b) Fall $\rightarrow$ Winter & (c) Winter & (d) Winter $\rightarrow$ Fall
  \end{tabular}
  \caption{
    \textbf{Cross Season Image Translation.} Example image-space conversions for the SYNTHIA seasons adaptation setting.
    We show real samples from each domain (Fall and Winter) alongside conversions to the opposite domain.
    }
  \label{fig:synthia}
\end{figure*}

%% file: tables/synthia.tex
\begin{table*}
  \centering
  \scriptsize
  \setlength{\tabcolsep}{3.0pt}
  \begin{tabular}{l|cccccccccccccggg}
    \toprule
    \multicolumn{17}{c}{\textbf{SYNTHIA Fall $\rightarrow$ Winter}} \\
    \midrule
    & \rot{sky} & \rot{building} & \rot{road} & \rot{sidewalk} & \rot{fence} & \rot{vegetation} & \rot{pole} & \rot{car} & \rot{traffic sign} & \rot{pedestrian} & \rot{bicycle} & \rot{lanemarking} & \rot{traffic light} & \rot{\textbf{mIoU}} & \rot{\textbf{fwIoU}} & \rot{\textbf{Pixel acc.}} \\ \midrule
    Source only            & 91.7 & 80.6 & 79.7 & 12.1 & 71.8 & 44.2 & 26.1 & 42.8 & 49.0 & 38.7 & 45.1 & 41.3 & 24.5 & 49.8 & 71.7 & 82.3 \\
    FCNs in the wild & 92.1 & 86.7 & 91.3 & 20.8 & 72.7 & \textbf{52.9} & \textbf{46.5} & 64.3 & \textbf{50.0} & \textbf{59.5} & \textbf{54.6} & \textbf{57.5} & \textbf{26.1} & 59.6 &  --- &  --- \\
    CyCADA pixel-only    & \textbf{92.5} & \textbf{90.1} & \textbf{91.9} & \textbf{79.9} & \textbf{85.7} & 47.1 & 36.9 & \textbf{82.6} & 45.0 & 49.1 & 46.2 & 54.6 & 21.5 & \textbf{63.3} & \textbf{85.7} & \textbf{92.1} \\ \midrule
    Oracle (Train on target)        & 93.8 & 92.2 & 94.7 & 90.7 & 90.2 & 64.4 & 38.1 & 88.5 & 55.4 & 51.0 & 52.0 & 68.9 & 37.3 & 70.5 & 89.9 & 94.5 \\
    \bottomrule
  \end{tabular}
  \caption{
    Adaptation between seasons in the SYNTHIA dataset. We report IoU for each class and mean IoU, freq-weighted IoU and pixel accuracy.
    Our CyCADA method achieves state-of-the-art performance on average across all categories.  $^*$FCNs in the wild is by \citet{hoffman_arxiv16}.
    }
  \label{table:synthia}
\end{table*}

%% file: figs/gta-cityscapes-seg.tex
\begin{figure}
  \centerline{
    \setlength{\tabcolsep}{2.0pt}
  \begin{tabular}{cccc}
    \includegraphics[width=\myw, height=\myh]{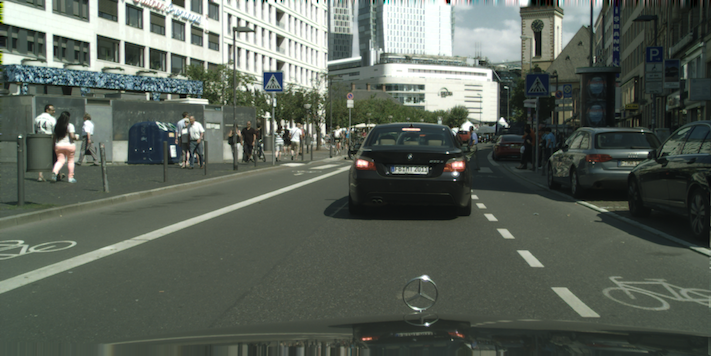} &
	\includegraphics[width=\myw, height=\myh]{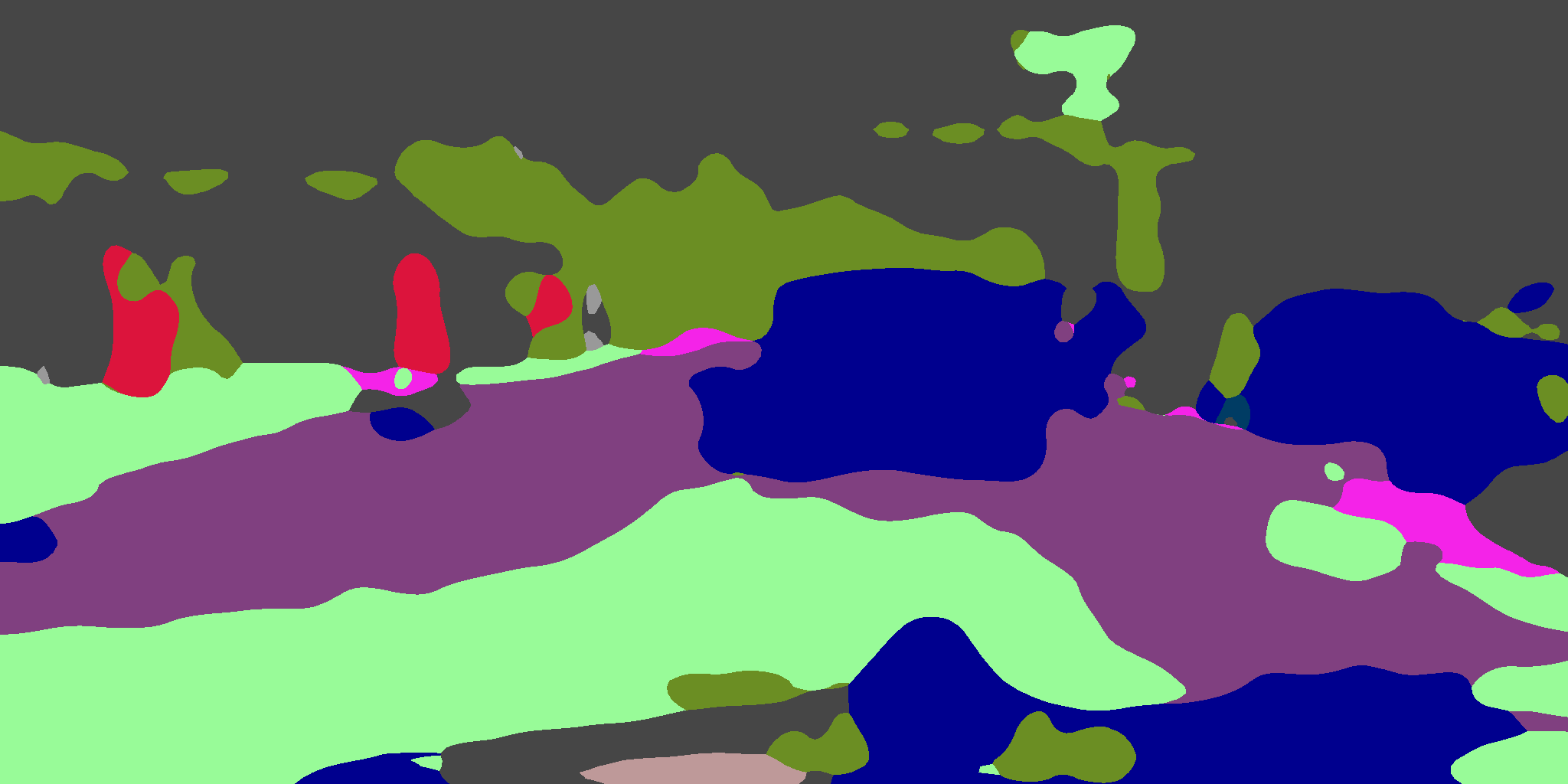} &
	\includegraphics[width=\myw, height=\myh]{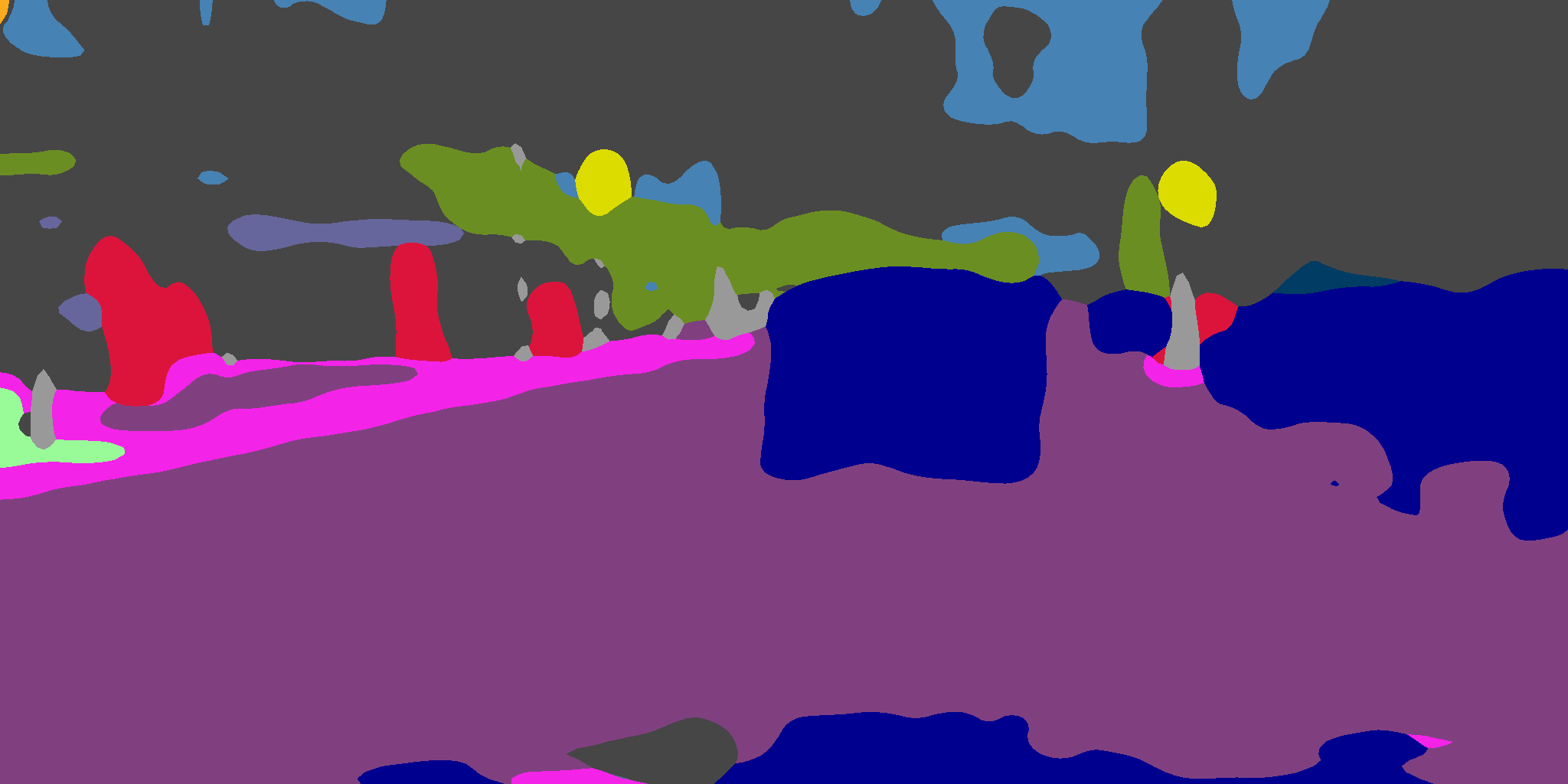} &
	\includegraphics[width=\myw, height=\myh]{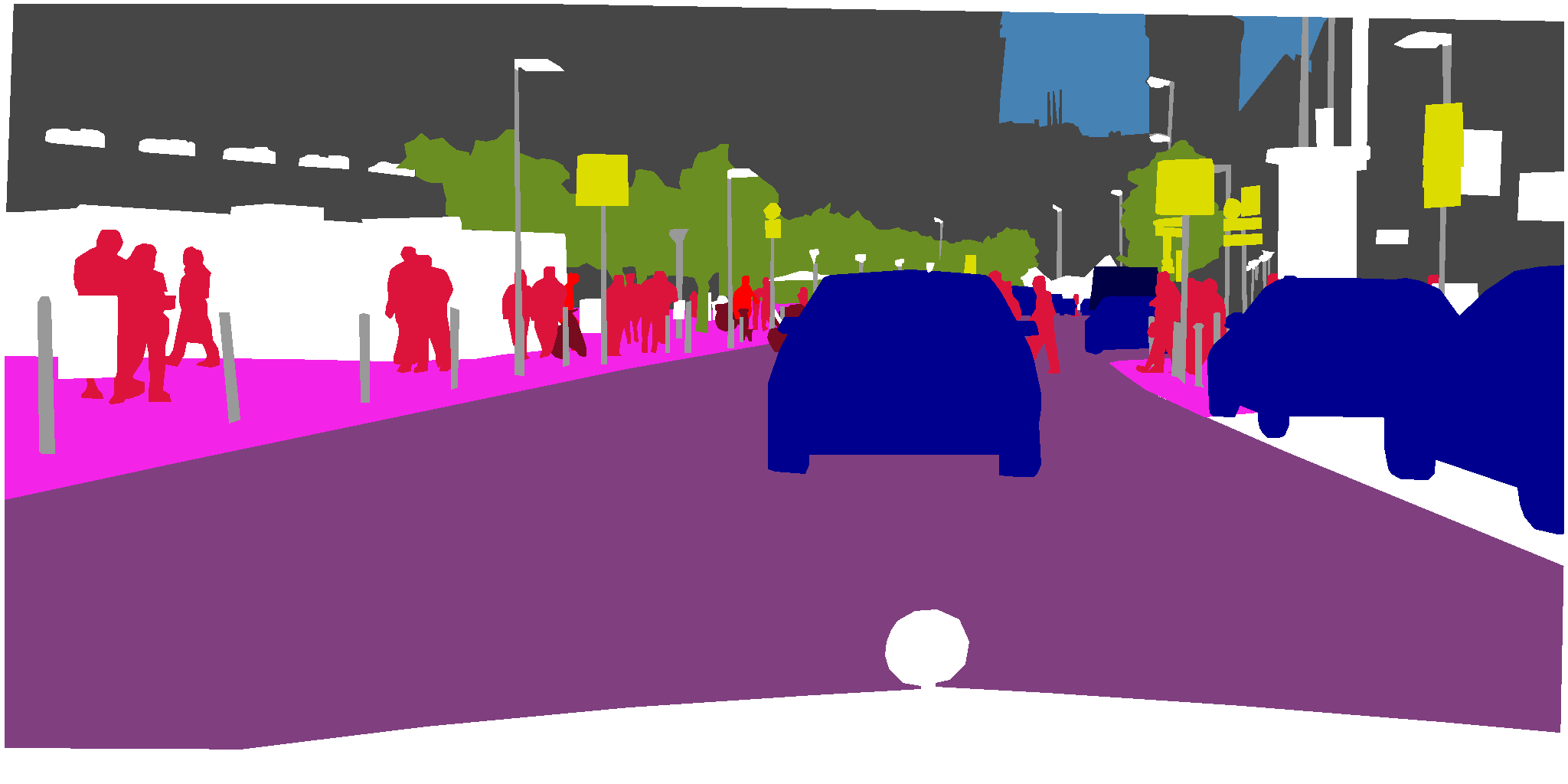} 
   \\
   \includegraphics[width=\myw, height=\myh]{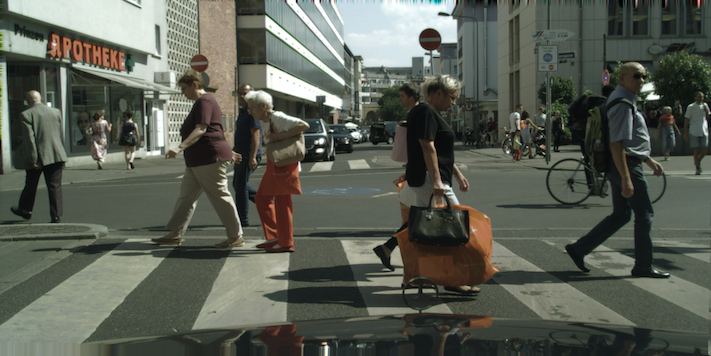} &
\includegraphics[width=\myw, height=\myh]{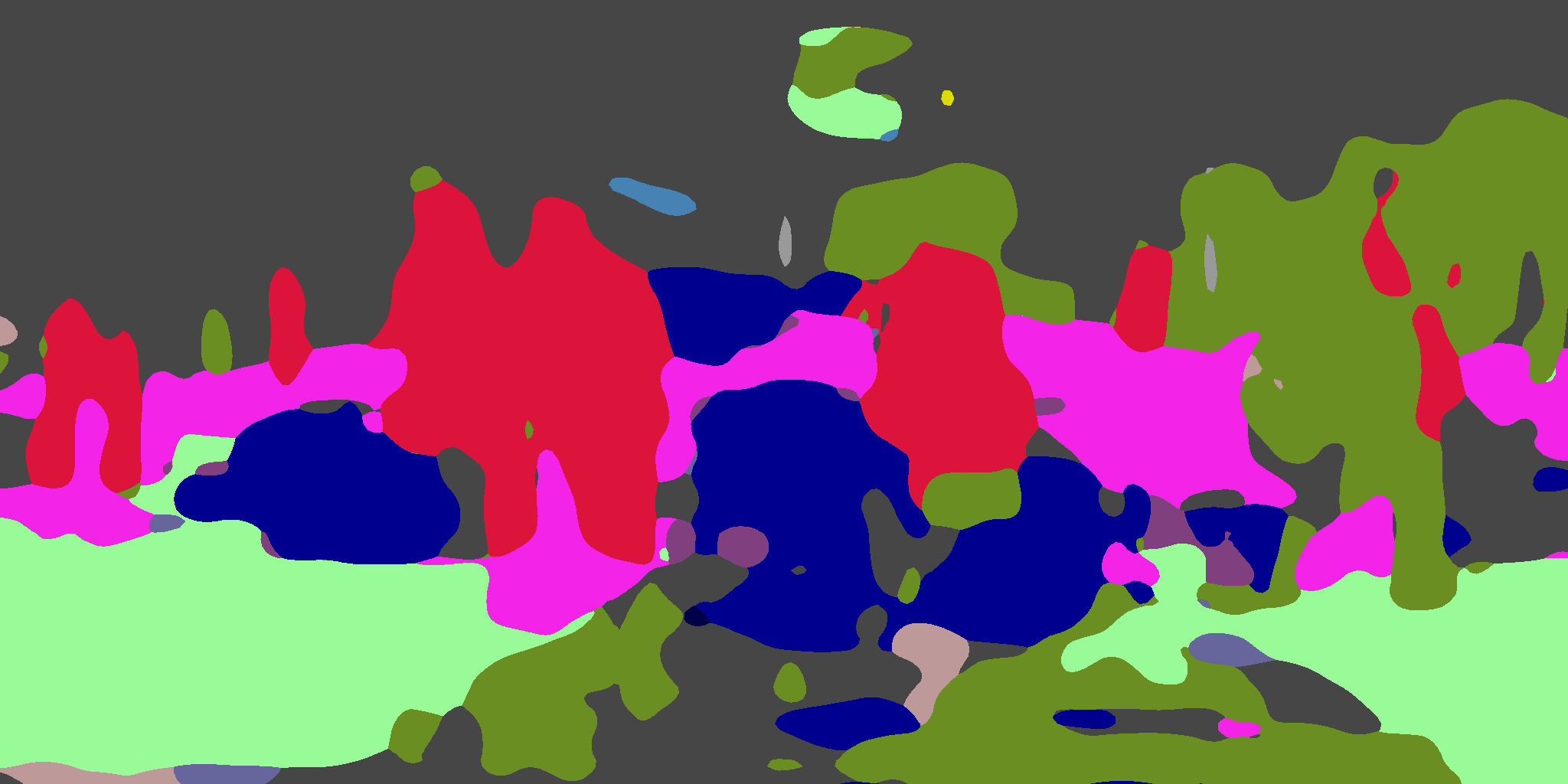} &
\includegraphics[width=\myw, height=\myh]{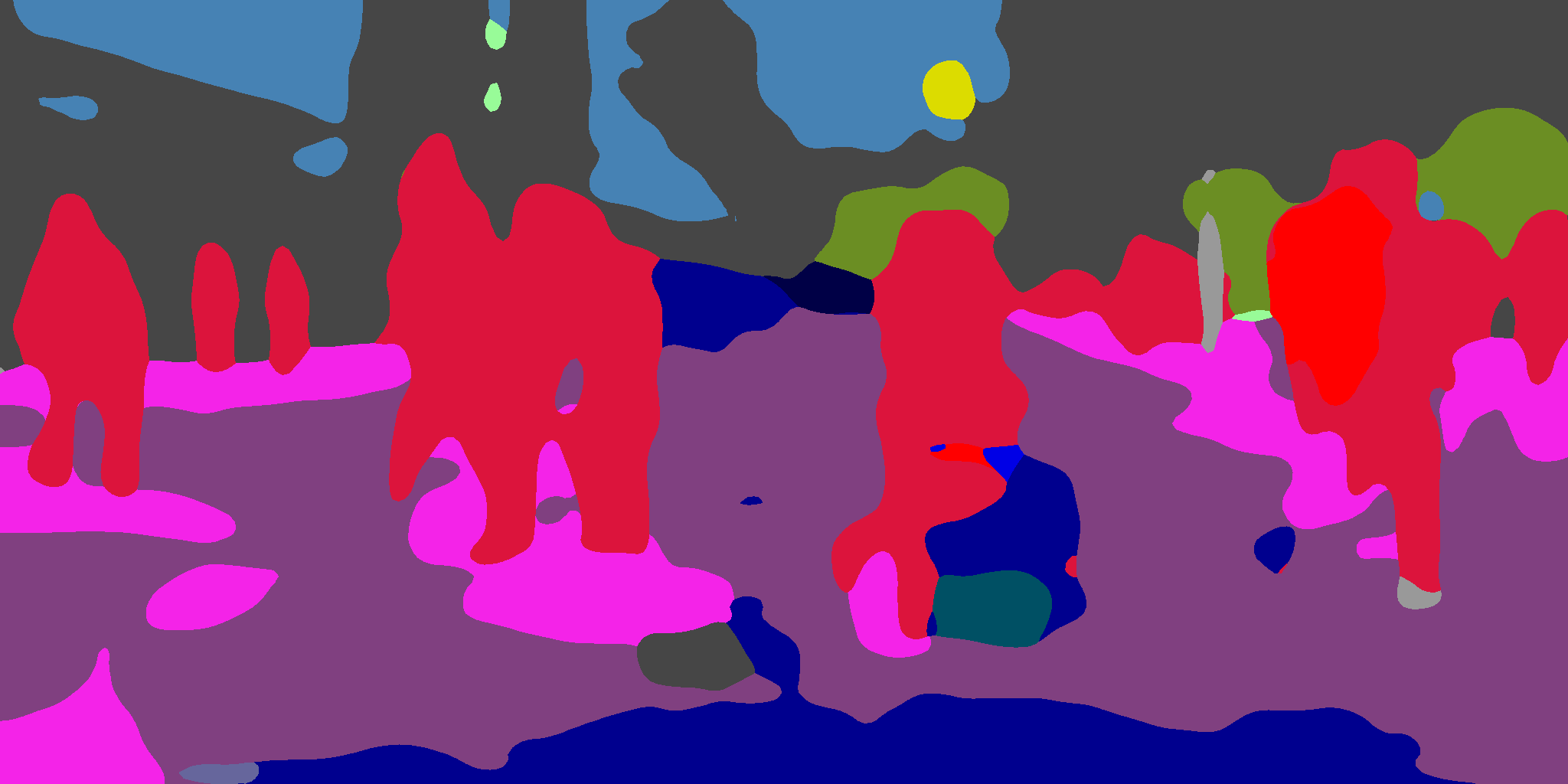} &
\includegraphics[width=\myw, height=\myh]{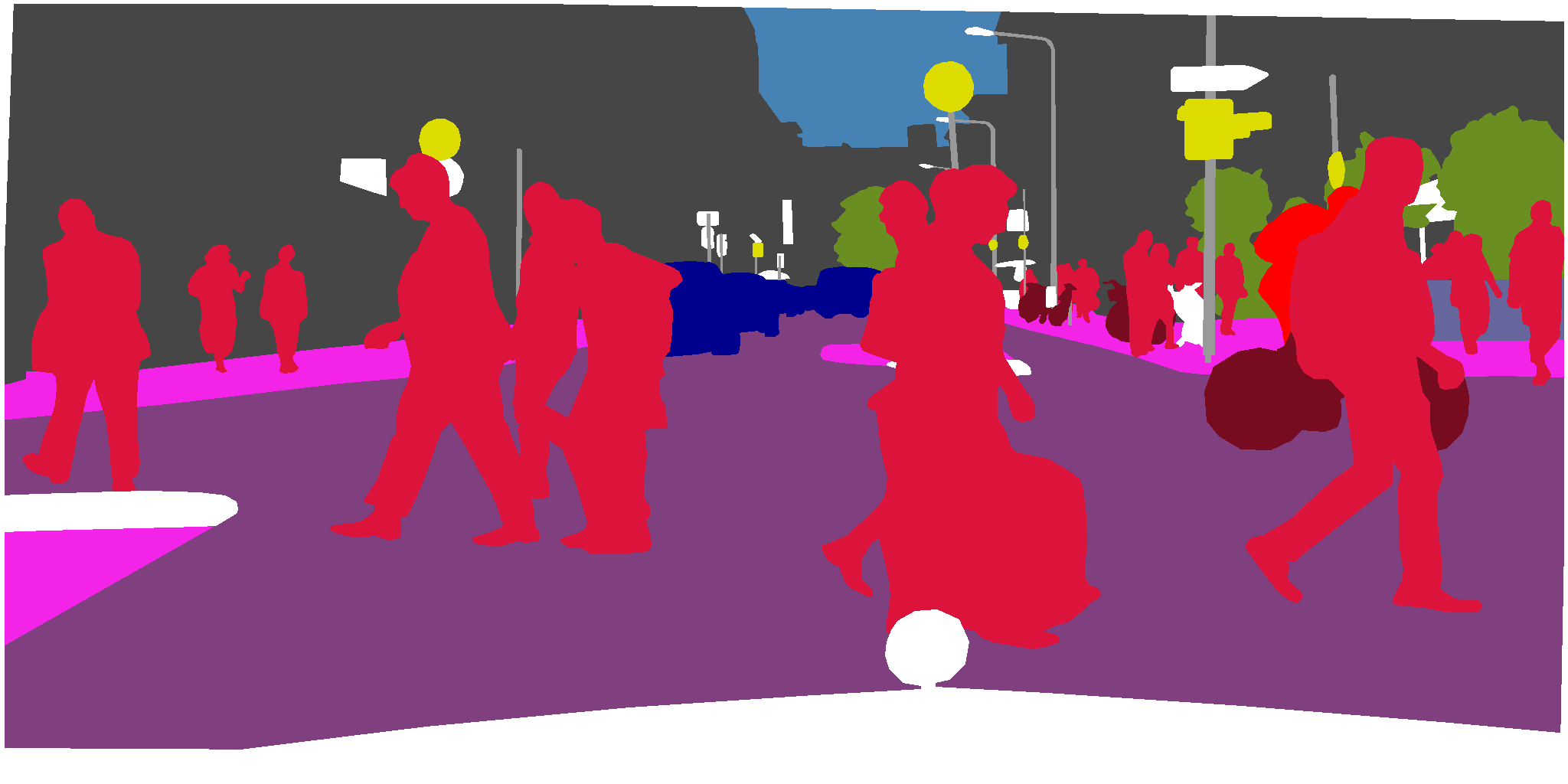}
\\
   (a) Test Image & (b) Source Prediction & (c) CyCADA Prediction & (d) Ground Truth\\
  \end{tabular}
  }
  \caption{
  \textbf{GTA5 to CityScapes Semantic Segmentation.} Each test CityScapes image (a) along with the corresponding predictions from the source only model (b) and our CyCADA model (c) are shown and may be compared against the ground truth annotation (d). }
  \label{fig:gta-cityscapes-seg}
\end{figure}

%% file: tables/gta-cityscapes.tex
\begin{table*}[h]
  \begin{center}
  \scriptsize
  \setlength{\tabcolsep}{1.35pt}
  \begin{tabular}{l|c|cccccccccccccccccccggg}
    \toprule
    \multicolumn{24}{c}{\textbf{GTA5 $\rightarrow$ Cityscapes}} \\
    \midrule
     &\rot{Architecture} & \rot{road} & \rot{sidewalk} & \rot{building} & \rot{wall} & \rot{fence} & \rot{pole} & \rot{traffic light} & \rot{traffic sign} & \rot{vegetation} & \rot{terrain} & \rot{sky} & \rot{person} & \rot{rider} & \rot{car} & \rot{truck} & \rot{bus} & \rot{train} & \rot{motorbike} & \rot{bicycle} & \rot{\textbf{mIoU}} & \rot{\textbf{fwIoU}} & \rot{\textbf{Pixel acc.}} \\ \midrule
    Source only & A            & 26.0 & 14.9 & 65.1 &  5.5 & 12.9 &  8.9 &  6.0 &  2.5 & 70.0 &  2.9 & 47.0 & 24.5 &  0.0 & 40.0 & 12.1 &  1.5 &  0.0 &  0.0 &  0.0 & 17.9 & 41.9 & 54.0 \\
    FCNs in the wild$^*$ & A & 70.4 & 32.4 & 62.1 & 14.9 &  5.4 & 10.9 & 14.2 &  2.7 & 79.2 & 21.3 & 64.6 & 44.1 &  4.2 & 70.4 &  8.0 &  7.3 &  0.0 &  3.5 &  0.0 & 27.1 & ---  & ---  \\
	CyCADA feat-only & A & \bf 85.6 &30.7& 74.7 &14.4 &13.0 &17.6 &13.7 &5.8 &74.6& 15.8 &\textbf{69.9} &38.2 &3.5 &72.3& 16.0& 5.0& 0.1 &3.6 &0.0& 29.2 & 71.5 & 82.5\\ 
    CyCADA pixel-only & A & {83.5} & \textbf{38.3} & \textbf{76.4} & {20.6} & \textbf{16.5} &{22.2} & \textbf{26.2} & \textbf{21.9} & \textbf{80.4} & {28.7} & {65.7} & {49.4} &  4.2 & {74.6} & {16.0} & {26.6} &  {2.0} &  {8.0} &  0.0 & {34.8} & {73.1} & {82.8} \\
	CyCADA pixel+feat & A & 85.2&	 37.2 	&\textbf{76.5} 	&\textbf{21.8} 	&15.0& 	\textbf{23.8}&	 22.9& 	21.5 	&\textbf{80.5} 	&\textbf{31.3} 	&60.7 	&\textbf{50.5} 	&\textbf{9.0}	 &\textbf{76.9} 	&\textbf{17.1} 	&\textbf{28.2} 	&\textbf{4.5} 	&\textbf{9.8} 	&0.0& \textbf{35.4} & \bf73.8 & \bf83.6\\ 
	\midrule
		Oracle - Target Super & A   & 96.4 & 74.5 & 87.1 & 35.3 & 37.8 & 36.4 & 46.9 & 60.1 & 89.0 & 54.3 & 89.8 & 65.6 & 35.9 & 89.4 & 38.6 & 64.1 & 38.6 & 40.5 & 65.1 & 60.3 & 87.6 & 93.1 \\
		\midrule
		\midrule
			Source only & B & 	42.7 &	26.3 &	51.7 &	5.5 &	6.8 &	13.8 &	23.6 &	6.9 &	75.5 &	11.5 &	36.8 &	49.3 &	0.9 &	46.7 &	3.4 &	5.0 &	0.0 &	5.0 &	1.4 & 21.7 &	47.4 &	62.5 \\
		CyCADA feat-only & B & 	78.1 &	31.1 &	71.2 &	10.3 &	14.1 &	29.8 &	28.1 &	20.9 &	74.0 &	16.8 &	51.9 &	53.6 &	6.1 &	65.4 &	8.2 &	20.9 &	1.8 &	13.9 &	5.9 & 31.7 &	67.4 &	78.4 \\
		 %
		CyCADA pixel-only & 	B & 63.7 &	24.7 &	69.3 &	21.2 &	17.0 &	30.3 &	33.0 &	
		\bf 32.0 &	80.5 &	25.3 &	62.3 &	62.0 &	\bf 15.1 &	73.1 &	19.8 &	23.6 &	5.5 &	16.2 &	\bf 28.7 & 37.0 &	63.8 &	75.4 \\ 
		CyCADA pixel+feat & B & \bf 79.1&	\bf 33.1&	\bf 77.9&	\bf 23.4&	\bf 17.3&	\bf 32.1&	\bf 33.3&	31.8& \bf	81.5&	\bf 26.7&	\bf 69.0	&\bf 62.8&	14.7&	\bf 74.5	&\bf 20.9	&\bf 25.6	&\bf 6.9&	\bf 18.8&	20.4 & \bf 39.5	&\bf 72.4&	\bf 82.3\\ 
		\midrule
		Oracle - Target Super & B   & 97.3	&79.8&	88.6	&32.5&	48.2&	56.3&	63.6	&73.3	&89.0&	58.9&	93.0&	78.2&	55.2&	92.2&	45.0&	67.3&	39.6&	49.9&	73.6 & 67.4	& 89.6 & 94.3\\
    \bottomrule
  \end{tabular}
  \end{center}
  \caption{
    Adaptation between GTA5 and Cityscapes, showing IoU for each class and mean IoU, freq-weighted IoU and pixel accuracy.
    CyCADA significantly outperforms baselines, nearly closing the gap to the target-trained oracle on pixel accuracy. $^*$FCNs in the wild is by \citet{hoffman_arxiv16}. We compare our model using two base semantic segmentation architectures (A) VGG16-FCN8s~\citep{long_cvpr15} base network and (B) DRN-26~\citep{drn}.
    }
  \label{table:gta-cityscapes}
\end{table*}

%% file: figs/gta-cityscapes.tex
\begin{figure}[h]
	\centerline{
  \setlength{\tabcolsep}{2.0pt}
  \begin{tabular}{cc cc}
    \includegraphics[width=\myw, height=\myh]{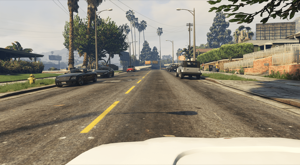} &
    \includegraphics[width=\myw, height=\myh]{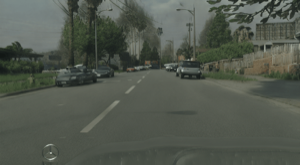} &
    \includegraphics[width=\myw, height=\myh]{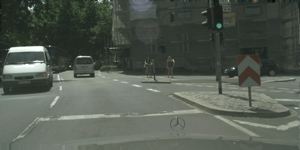} &
    \includegraphics[width=\myw, height=\myh]{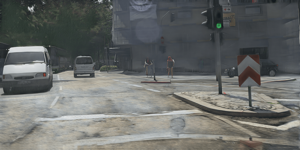}
    \\
    \includegraphics[width=\myw, height=\myh]{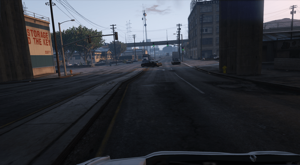} &
    \includegraphics[width=\myw, height=\myh]{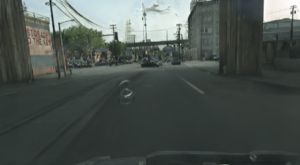} &
    \includegraphics[width=\myw, height=\myh]{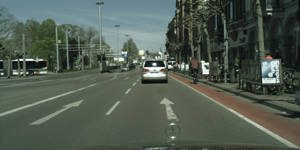} &
    \includegraphics[width=\myw, height=\myh]{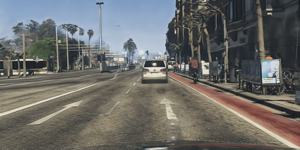}
   \\
   (a) GTA5 & (b) GTA5 $\rightarrow$ Cityscapes & (c) CityScapes & (d) CityScapes $\rightarrow$ GTA5
  \end{tabular}
  }
  \caption{
  \textbf{GTA5 to CityScapes Image Translation.} Example images from the GTA5 (a) and Cityscapes (c) datasets, alongside their image-space conversions to the opposite domain, (b) and (d), respectively. Our model achieves highly realistic domain conversions.}
  \label{fig:gta-cityscapes}
\end{figure}

%% file: conclusion.tex
\section{Conclusion}

We presented a cycle-consistent adversarial domain adaptation method that unifies cycle-consistent adversarial models with adversarial adaptation methods.
CyCADA is able to adapt even in the absence of target labels and is broadly applicable at both the pixel-level and in feature space.
An image-space adaptation instantiation of CyCADA also provides additional interpretability and serves as a useful way to verify successful adaptation.
Finally, we experimentally validated our model on a variety of adaptation tasks: state-of-the-art results in multiple evaluation settings indicate its effectiveness, even on challenging synthetic-to-real tasks. 

%% file: references.tex
\bibliography{references}
\bibliographystyle{iclr2018_conference}

%% file: appendix.tex
\pagebreak
\section{Appendix}

\subsection{Implementation Details}
We begin by pretraining the source task model, $f_S$, using the task loss on the labeled source data. Next, we perform pixel-level adaptation using our image space GAN losses together with semantic consistency and cycle consistency losses. This yeilds learned parameters for the image transformations, $\StoT$ and $\TtoS$, image discriminators, $D_S$ and $D_T$, as well as an initial setting of the task model, $f_T$, which is trained using pixel transformed source images and the corresponding source pixel labels. Finally, we perform feature space adpatation in order to update the target semantic model, $f_T$, to have features which are aligned between the source images mapped into target style and the real target images. During this phase, we learn the feature discriminator, $D_\text{feat}$ and use this to guide the representation update to $f_T$. In general, our method could also perform phases 2 and 3 simultaneously, but this would require more GPU memory then available at the time of these experiments. 

For all feature space adaptation we equally weight the generator and discriminator losses. We only update the generator when the discriminator accuracy is above 60\% over the last batch (digits) or last 100 iterations (semantic segmentation) -- this reduces the potential for volatile training. If after an epoch (entire pass over dataset) no suitable discriminator is found, the feature adaptation stops, otherwise it continues until max iterations are reached.

\subsubsection{Digit Experiments}
\label{sec:digit-details}
\begin{figure}
	\includegraphics[width=.24\linewidth]{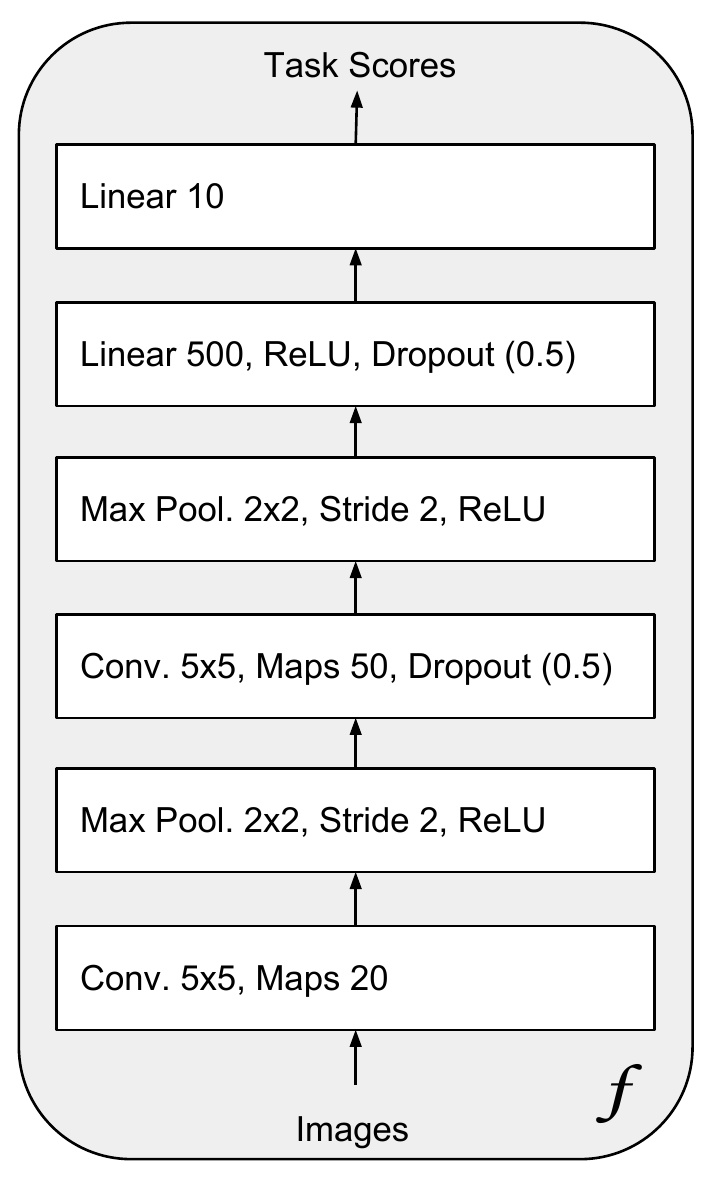}
	\includegraphics[width=.24\linewidth]{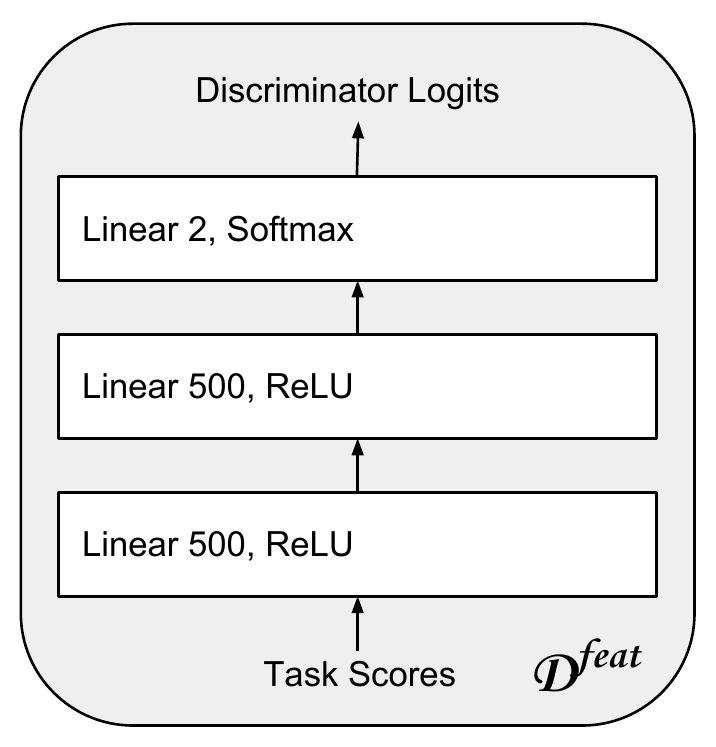}
	\includegraphics[width=.24\linewidth]{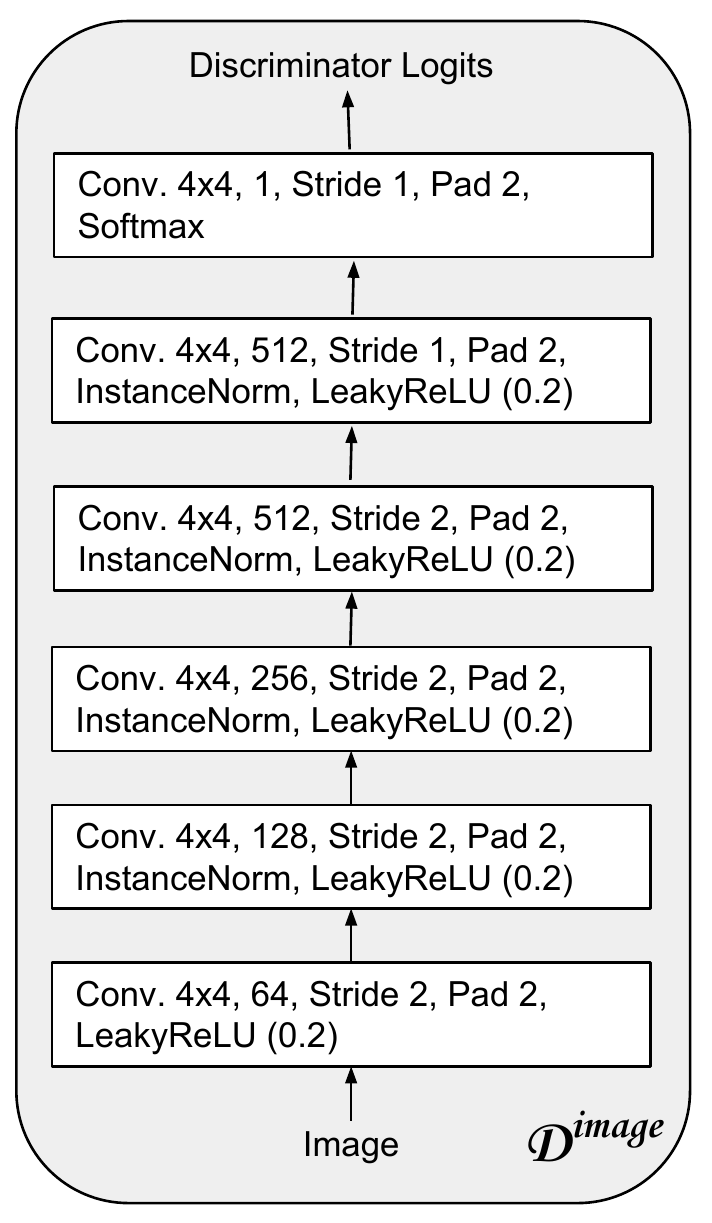}
	\includegraphics[width=.24\linewidth]{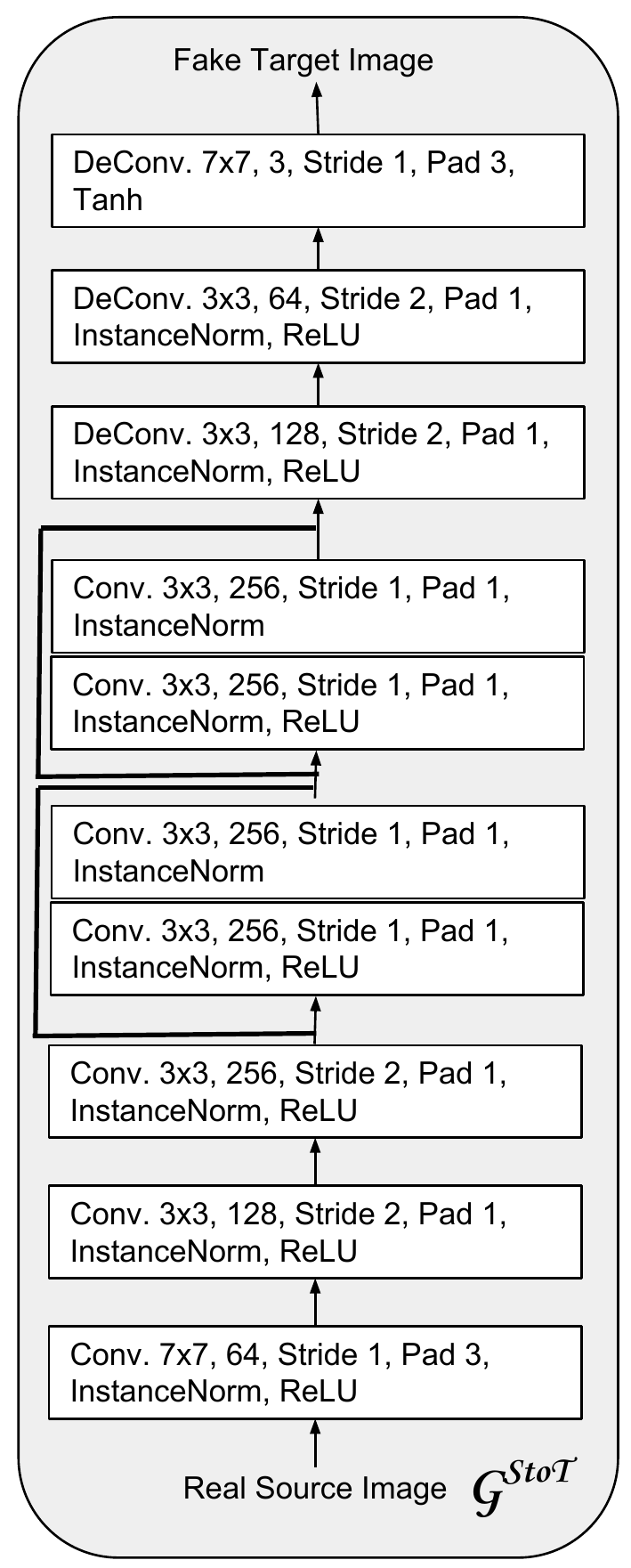}
	\caption{Network architectures used for digit experiments. We show here the task net ($f$), discriminator for feature level adaptation ($D^{feat}$), discriminator for image level adaptation ($D^{image}$), and generator for source to target ($G$) -- same network used for target to source.}
	\label{fig:digit_nets}
\end{figure}
For all digit experiments we use a variant of the LeNet architecture as the task net (Figure~\ref{fig:digit_nets} \textit{left}). Our feature discriminator network consists of 3 fully connected layers (Figure~\ref{fig:digit_nets} \textit{mid left}). The image discriminator network consists of 6 convolutional layers culminating in a single value per pixel (Figure~\ref{fig:digit_nets} \textit{mid right}). Finally, to generate one image domain from another we use a multilayer network which consists of convolution layers followed by two residual blocks and then deconvolution layers (Figure~\ref{fig:digit_nets} \textit{right}). 
All stages are trained using the Adam optimizer. 

\paragraph{Hyperparameters.} For training the source task net model, we use learning rate 1e-4 and train for 100 epochs over the data with batch size 128. For feature space adaptation we use learning rate 1e-5 and train for max 200 epochs over the data. For pixel space adaptation we train our generators and discriminators with equal weighting on all losses, use batch size 100, learning rate 2e-4 (default from CycleGAN), and trained for 50 epochs. We ran each experiment 4 times and report the average and standard error across the runs.

\subsubsection{Semantic Segmentation}
\label{sec:ss-details}

\begin{figure}[h]
	\centerline{
  \setlength{\tabcolsep}{2.0pt}
  \begin{tabular}{cc cc}
    \includegraphics[width=\myw, height=\myh]{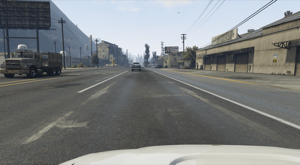} &
    \includegraphics[width=\myw, height=\myh]{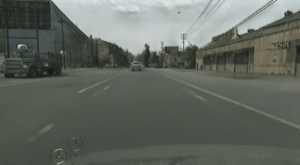} &
    \includegraphics[width=\myw, height=\myh]{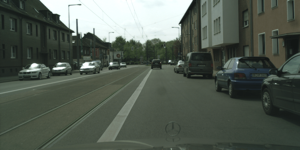} &
    \includegraphics[width=\myw, height=\myh]{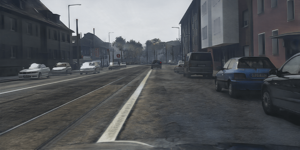}
   \\
    \includegraphics[width=\myw, height=\myh]{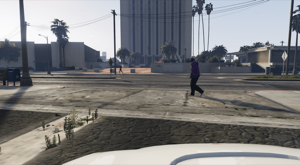} &
    \includegraphics[width=\myw, height=\myh]{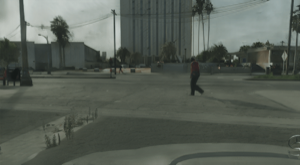} &
    \includegraphics[width=\myw, height=\myh]{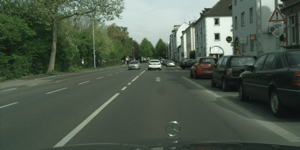} &
    \includegraphics[width=\myw, height=\myh]{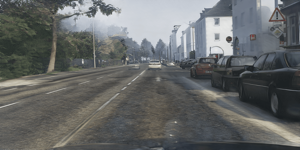}
   \\
    \includegraphics[width=\myw, height=\myh]{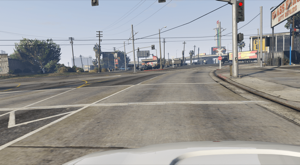} &
    \includegraphics[width=\myw, height=\myh]{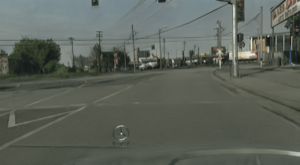} &
    \includegraphics[width=\myw, height=\myh]{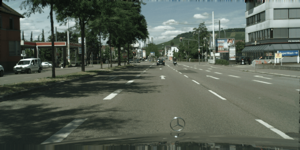} &
    \includegraphics[width=\myw, height=\myh]{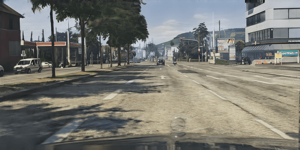}
   \\
    \includegraphics[width=\myw, height=\myh]{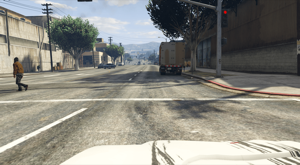} &
    \includegraphics[width=\myw, height=\myh]{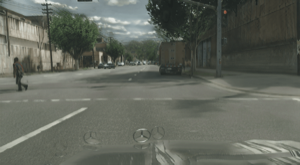} &
    \includegraphics[width=\myw, height=\myh]{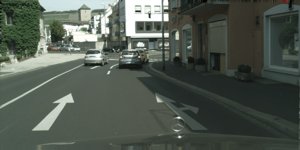} &
    \includegraphics[width=\myw, height=\myh]{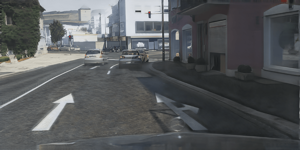}
   \\
    \includegraphics[width=\myw, height=\myh]{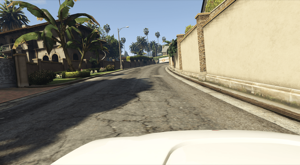} &
    \includegraphics[width=\myw, height=\myh]{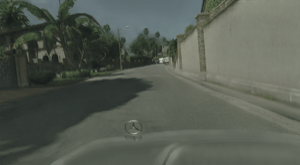} &
    \includegraphics[width=\myw, height=\myh]{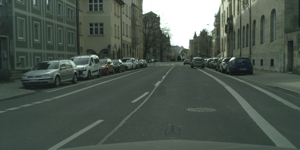} &
    \includegraphics[width=\myw, height=\myh]{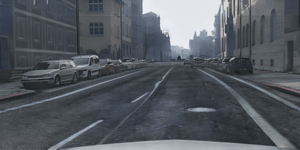}
   \\
    \includegraphics[width=\myw, height=\myh]{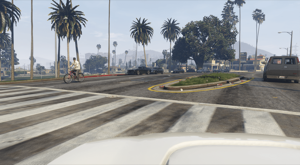} &
    \includegraphics[width=\myw, height=\myh]{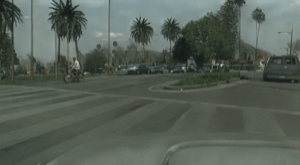} &
    \includegraphics[width=\myw, height=\myh]{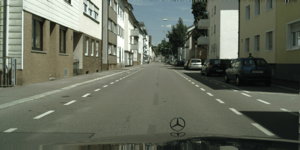} &
    \includegraphics[width=\myw, height=\myh]{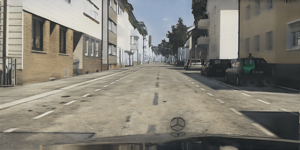}
   \\
   (a) GTA5 & (b) GTA5 $\rightarrow$ Cityscapes & (c) CityScapes & (d) CityScapes $\rightarrow$ GTA5
  \end{tabular}
  }
  \caption{
  \textbf{GTA5 to CityScapes Image Translation.} Example images from the GTA5 (a) and Cityscapes (c) datasets, alongside their image-space conversions to the opposite domain, (b) and (d), respectively. Our model achieves highly realistic domain conversions.}
  \label{fig:gta-cityscapes-appendix}
\end{figure}

We experiment with both the VGG16-FCN8s~\cite{long_cvpr15} architecture as well as the DRN-26~\cite{drn} architecture. 
For FCN8s, we train our source semantic segmentation model for 100k iterations using SGD with learning rate 1e-3 and momentum 0.9.
For the DRN-26 architecture, we train our source semantic segmentation model for 115K iterations using SGD with learning rate 1e-3 and momentum 0.9. We use a crop size of 600x600 and a batch size of 8 for this training. 
For cycle-consistent image level adaptation, we followed the network architecture and hyperparameters of CycleGAN\citep{zhu_arxiv17}.
All images were resized to have width of 1024 pixels while keeping the aspect ratio, and the training was performed with randomly cropped patches of size 400 by 400. Also, due to large size of the dataset, we trained only 20 epochs.
For feature level adaptation, we train using SGD with momentum, 0.99, and learning rate 1e-5. We weight the representation loss ten times less than the discriminator loss as a convienience since otherwise the discriminator did not learn a suitable model within a single epoch. Then the segmentation model was trained separately using the adapted source images and the ground truth labels of the source data. Due to memory limitations we can only include a single source and single target image at a time (crops of size 768x768), this small batch is one of the main reasons for using a high momentum parameter.

\subsection{Comparison to \citet{shrivastava_cvpr17} for Semantic Segmentation}
\begin{figure}
	\centering
	\begin{tabular}{ccc}
		\includegraphics[width=.32\linewidth]{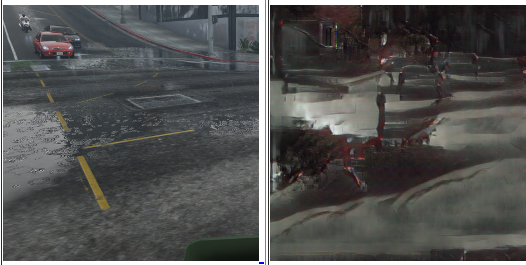} &
		\includegraphics[width=.32\linewidth]{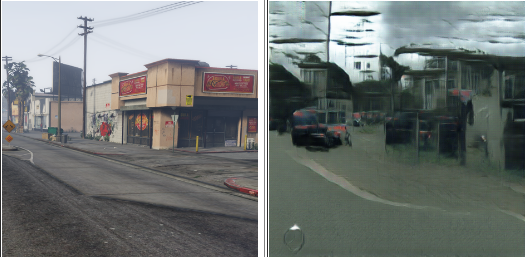}&
		\includegraphics[width=.32\linewidth]{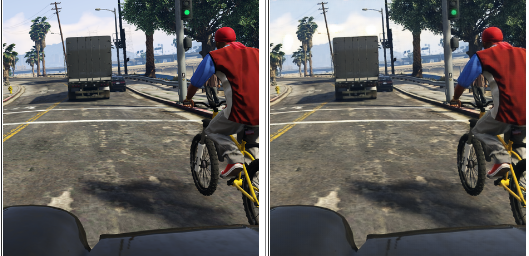}\\
		(a) $\lambda = 1$ & (b) $\lambda = 2.5$ & (c) $\lambda = 10$	
\end{tabular}

\caption{Image transformation results from \citet{shrivastava_cvpr17} applied to GTA to CityScapes transformation. We demonstrate results using three different settings for $\lambda$.}
\label{fig:shrivastava}
\end{figure}

We illustrate the performance of a recent pixel level adaptation approach proposed by \citet{shrivastava_cvpr17} on our semantic segmentation data -- GTA to Cityscapes. These images are significantly larger and more complex than those shown in the experiments in the original paper. We show image to image translation results under three different settings of the model hyperparameter, $\lambda$, which controls the tradeoff between the reconstruction loss and the visual style loss. When $\lambda=10$ (Figure~\ref{fig:shrivastava} \textit{right}), the resulting image converges to a near replica of the original image, thus preserving content but lacking the correct target style. When $\lambda=1$ or $\lambda=2.5$ (Figure~\ref{fig:shrivastava} \textit{left}), the results lack any consistent semantics making it difficult to perceive the style of the transformed image. Thus, the resulting performance for this model is 11.6 mIoU for FCN8s with VGG, well below the performance of the corresponding source model of 17.9 mIoU. 

\subsection{Experiment Analysis}
\begin{figure}
	\centering
	\begin{tabular}{cc}
		\includegraphics[width=.45\linewidth]{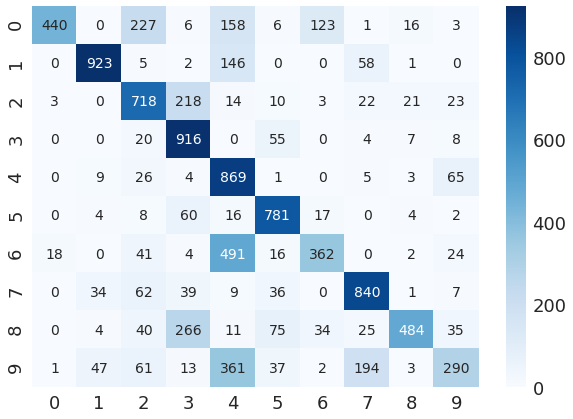} &
		\includegraphics[width=.45\linewidth]{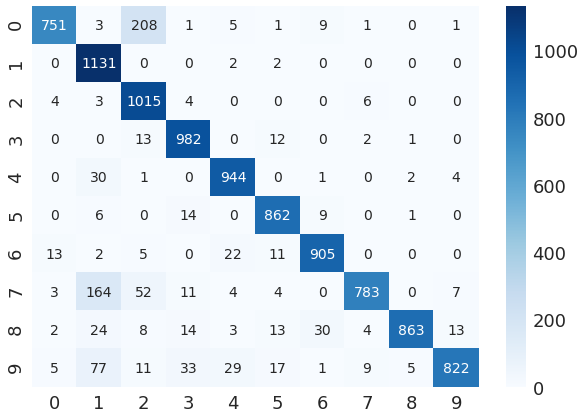}\\
		(a) Source only Model & (b) CyCADA model
	\end{tabular}	
	\caption{Confusion matricies for SVHN $\rightarrow$ MNIST experiment. }
	\label{fig:digit_err}
\end{figure}

To understand the types of mistakes which are improved upon and those which still persist after adaptation, we present the confusion matricies before and after our approach for the digit experiment of SVHN to MNIST (Figure~\ref{fig:digit_err}). Before adaptation we see common confusions are 0s with 2s, 4s, and 7s. 6 with 4, 8 with 3, and 9 with 4. After adaptation all errors are reduced, but we still find that 7s are confused with 1s and 0s with 2s. These errors make some sense as with hand written digits, these digits sometimes resemble one another. It remains an open question to produce a model which may overcome these types of errors between highly similar classes.  

%% file: 2018-iclr-cycle-adaptation.bbl
\begin{thebibliography}{49}
\providecommand{\natexlab}[1]{#1}
\providecommand{\url}[1]{\texttt{#1}}
\expandafter\ifx\csname urlstyle\endcsname\relax
  \providecommand{\doi}[1]{doi: #1}\else
  \providecommand{\doi}{doi: \begingroup \urlstyle{rm}\Url}\fi

\bibitem[Arjovsky et~al.(2017)Arjovsky, Chintala, and Bottou]{arjovsky_arxiv17}
Martin Arjovsky, Soumith Chintala, and Léon Bottou.
\newblock Wasserstein gan.
\newblock \emph{CoRR}, abs/1701.07875, 2017.
\newblock URL \url{http://arxiv.org/abs/1701.07875}.

\bibitem[Bousmalis et~al.(2017{\natexlab{a}})Bousmalis, Irpan, Wohlhart, Bai,
  Kelcey, Kalakrishnan, Downs, Ibarz, Pastor, Konolige, Levine, and
  Vanhoucke]{bousmalis_arxiv17_robotic}
Konstantinos Bousmalis, Alex Irpan, Paul Wohlhart, Yunfei Bai, Matthew Kelcey,
  Mrinal Kalakrishnan, Laura Downs, Julian Ibarz, Peter Pastor, Kurt Konolige,
  Sergey Levine, and Vincent Vanhoucke.
\newblock Using simulation and domain adaptation to improve efficiency of deep
  robotic grasping.
\newblock \emph{CoRR}, abs/1709.07857, 2017{\natexlab{a}}.
\newblock URL \url{http://arxiv.org/abs/1709.07857}.

\bibitem[Bousmalis et~al.(2017{\natexlab{b}})Bousmalis, Silberman, Dohan,
  Erhan, and Krishnan]{bousmalis_cvpr17}
Konstantinos Bousmalis, Nathan Silberman, David Dohan, Dumitru Erhan, and Dilip
  Krishnan.
\newblock Unsupervised pixel-level domain adaptation with generative
  adversarial networks.
\newblock In \emph{The IEEE Conference on Computer Vision and Pattern
  Recognition (CVPR)}, July 2017{\natexlab{b}}.

\bibitem[Chen et~al.(2017)Chen, Chen, Chen, Tsai, Frank~Wang, and
  Sun]{chen_iccv17}
Yi-Hsin Chen, Wei-Yu Chen, Yu-Ting Chen, Bo-Cheng Tsai, Yu-Chiang Frank~Wang,
  and Min Sun.
\newblock No more discrimination: Cross city adaptation of road scene
  segmenters.
\newblock In \emph{The IEEE International Conference on Computer Vision
  (ICCV)}, Oct 2017.

\bibitem[Cordts et~al.(2016)Cordts, Omran, Ramos, Rehfeld, Enzweiler, Benenson,
  Franke, Roth, and Schiele]{cordts_cvpr16}
Marius Cordts, Mohamed Omran, Sebastian Ramos, Timo Rehfeld, Markus Enzweiler,
  Rodrigo Benenson, Uwe Franke, Stefan Roth, and Bernt Schiele.
\newblock The {Cityscapes} dataset for semantic urban scene understanding.
\newblock In \emph{Proc. of the IEEE Conference on Computer Vision and Pattern
  Recognition (CVPR)}, 2016.

\bibitem[Denton et~al.(2015)Denton, Chintala, Fergus, et~al.]{denton2015deep}
Emily~L Denton, Soumith Chintala, Rob Fergus, et~al.
\newblock Deep generative image models using a laplacian pyramid of adversarial
  networks.
\newblock In \emph{Neural Information Processing Systems ({NIPS})}, pp.\
  1486--1494, 2015.

\bibitem[Donahue et~al.(2017)Donahue, Kr{\"a}henb{\"u}hl, and
  Darrell]{donahue2016adversarial}
Jeff Donahue, Philipp Kr{\"a}henb{\"u}hl, and Trevor Darrell.
\newblock Adversarial feature learning.
\newblock In \emph{ICLR}, 2017.

\bibitem[Dumoulin et~al.(2016)Dumoulin, Belghazi, Poole, Lamb, Arjovsky,
  Mastropietro, and Courville]{dumoulin2016adversarially}
Vincent Dumoulin, Ishmael Belghazi, Ben Poole, Alex Lamb, Martin Arjovsky,
  Olivier Mastropietro, and Aaron Courville.
\newblock Adversarially learned inference.
\newblock \emph{arXiv preprint arXiv:1606.00704}, 2016.

\bibitem[Ganin et~al.(2016)Ganin, Ustinova, Ajakan, Germain, Larochelle,
  Laviolette, Marchand, and Lempitsky]{dann}
Y.~Ganin, E.~Ustinova, H.~Ajakan, P.~Germain, H.~Larochelle, F.~Laviolette,
  M.~Marchand, and V.~Lempitsky.
\newblock Domain-adversarial training of neural networks.
\newblock \emph{Journal of Machine Learning Research}, 2016.

\bibitem[Ganin \& Lempitsky(2015)Ganin and Lempitsky]{ganin_icml15}
Yaroslav Ganin and Victor Lempitsky.
\newblock Unsupervised domain adaptation by backpropagation.
\newblock In David Blei and Francis Bach (eds.), \emph{Proceedings of the 32nd
  International Conference on Machine Learning (ICML-15)}, pp.\  1180--1189.
  JMLR Workshop and Conference Proceedings, 2015.
\newblock URL \url{http://jmlr.org/proceedings/papers/v37/ganin15.pdf}.

\bibitem[Gatys et~al.(2016)Gatys, Ecker, and Bethge]{gatys2016image}
Leon~A Gatys, Alexander~S Ecker, and Matthias Bethge.
\newblock Image style transfer using convolutional neural networks.
\newblock In \emph{Computer Vision and Pattern Recognition ({CVPR})}, pp.\
  2414--2423, 2016.

\bibitem[Ghifary et~al.(2016)Ghifary, Kleijn, Zhang, Balduzzi, and
  Li]{ghifary_eccv16}
Muhammad Ghifary, W~Bastiaan Kleijn, Mengjie Zhang, David Balduzzi, and Wen Li.
\newblock Deep reconstruction-classification networks for unsupervised domain
  adaptation.
\newblock In \emph{European Conference on Computer Vision (ECCV)}, pp.\
  597--613. Springer, 2016.

\bibitem[Goodfellow et~al.(2014)Goodfellow, Pouget-Abadie, Mirza, Xu,
  Warde-Farley, Ozair, Courville, and Bengio]{goodfellow_nips14}
Ian Goodfellow, Jean Pouget-Abadie, Mehdi Mirza, Bing Xu, David Warde-Farley,
  Sherjil Ozair, Aaron Courville, and Yoshua Bengio.
\newblock Generative adversarial nets.
\newblock In \emph{Advances in Neural Information Processing Systems 27}. 2014.
\newblock URL
  \url{http://papers.nips.cc/paper/5423-generative-adversarial-nets.pdf}.

\bibitem[Hinton et~al.(2015)Hinton, Vinyals, and Dean]{hinton_arxiv15}
G.~Hinton, O.~Vinyals, and J.~Dean.
\newblock Distilling the knowledge in a neural network.
\newblock 2015.

\bibitem[Hoffman et~al.(2016)Hoffman, Wang, Yu, and Darrell]{hoffman_arxiv16}
Judy Hoffman, Dequan Wang, Fisher Yu, and Trevor Darrell.
\newblock {FCN}s in the wild: Pixel-level adversarial and constraint-based
  adaptation.
\newblock \emph{CoRR}, abs/1612.02649, 2016.
\newblock URL \url{http://arxiv.org/abs/1612.02649}.

\bibitem[Isola et~al.(2016)Isola, Zhu, Zhou, and Efros]{isola2016image}
Phillip Isola, Jun-Yan Zhu, Tinghui Zhou, and Alexei~A Efros.
\newblock Image-to-image translation with conditional adversarial networks.
\newblock \emph{arXiv preprint arXiv:1611.07004}, 2016.

\bibitem[Karacan et~al.(2016)Karacan, Akata, Erdem, and
  Erdem]{karacan2016learning}
Levent Karacan, Zeynep Akata, Aykut Erdem, and Erkut Erdem.
\newblock Learning to generate images of outdoor scenes from attributes and
  semantic layouts.
\newblock \emph{arXiv preprint arXiv:1612.00215}, 2016.

\bibitem[Kim et~al.(2017)Kim, Cha, Kim, Lee, and Kim]{kim_arxiv17}
Taeksoo Kim, Moonsu Cha, Hyunsoo Kim, Jung~Kwon Lee, and Jiwon Kim.
\newblock Learning to discover cross-domain relations with generative
  adversarial networks.
\newblock \emph{arXiv preprint arXiv:1703.05192}, 2017.

\bibitem[LeCun et~al.(1998)LeCun, Bottou, Bengio, and Haffner]{lecun_ieee98}
Y.~LeCun, L.~Bottou, Y.~Bengio, and P.~Haffner.
\newblock Gradient-based learning applied to document recognition.
\newblock \emph{Proceedings of the IEEE}, 86\penalty0 (11):\penalty0
  2278--2324, November 1998.

\bibitem[Levinkov \& Fritz(2013)Levinkov and Fritz]{levinkov_iccv13}
Evgeny Levinkov and Mario Fritz.
\newblock Sequential bayesian model update under structured scene prior for
  semantic road scenes labeling.
\newblock In \emph{IEEE International Conference on Computer Vision (ICCV)},
  2013.
\newblock URL
  \url{http://scalable.mpi-inf.mpg.de/files/2013/10/levinkov13iccv.pdf
  http://www.d2.mpi-inf.mpg.de/sequential-bayesian-update}.

\bibitem[Liu \& Tuzel(2016{\natexlab{a}})Liu and Tuzel]{cogan}
M.-Y. Liu and O.~Tuzel.
\newblock Coupled generative adversarial networks.
\newblock In \emph{Advances in Neural Information Processing Systems},
  2016{\natexlab{a}}.

\bibitem[Liu \& Tuzel(2016{\natexlab{b}})Liu and Tuzel]{liu_arxiv16}
Ming{-}Yu Liu and Oncel Tuzel.
\newblock Coupled generative adversarial networks.
\newblock In \emph{Neural Information Processing Systems ({NIPS})},
  2016{\natexlab{b}}.

\bibitem[Long et~al.(2015)Long, Shelhamer, and Darrell]{long_cvpr15}
Jonathan Long, Evan Shelhamer, and Trevor Darrell.
\newblock Fully convolutional networks for semantic segmentation.
\newblock \emph{CVPR (to appear)}, November 2015.

\bibitem[Long \& Wang(2015)Long and Wang]{long_icml15}
Mingsheng Long and Jianmin Wang.
\newblock Learning transferable features with deep adaptation networks.
\newblock In \emph{International Conference on Machine Learning ({ICML})},
  2015.

\bibitem[Mirza \& Osindero(2014)Mirza and Osindero]{mirza_arxiv14}
Mehdi Mirza and Simon Osindero.
\newblock Conditional generative adversarial nets.
\newblock \emph{CoRR}, abs/1411.1784, 2014.
\newblock URL \url{http://arxiv.org/abs/1411.1784}.

\bibitem[Netzer et~al.(2011)Netzer, Wang, Coates, Bissacco, Wu, and
  Ng]{netzer_nips11}
Yuval Netzer, Tao Wang, Adam Coates, Alessandro Bissacco, Bo~Wu, and Andrew~Y.
  Ng.
\newblock Reading digits in natural images with unsupervised feature learning.
\newblock In \emph{NIPS Workshop on Deep Learning and Unsupervised Feature
  Learning 2011}, 2011.
\newblock URL
  \url{http://ufldl.stanford.edu/housenumbers/nips2011_housenumbers.pdf}.

\bibitem[Radford et~al.(2015)Radford, Metz, and
  Chintala]{radford2015unsupervised}
Alec Radford, Luke Metz, and Soumith Chintala.
\newblock Unsupervised representation learning with deep convolutional
  generative adversarial networks.
\newblock \emph{arXiv preprint arXiv:1511.06434}, 2015.

\bibitem[Richter et~al.(2016)Richter, Vineet, Roth, and Koltun]{richter_eccv16}
Stephan~R. Richter, Vibhav Vineet, Stefan Roth, and Vladlen Koltun.
\newblock Playing for data: {G}round truth from computer games.
\newblock In Bastian Leibe, Jiri Matas, Nicu Sebe, and Max Welling (eds.),
  \emph{European Conference on Computer Vision (ECCV)}, volume 9906 of
  \emph{LNCS}, pp.\  102--118. Springer International Publishing, 2016.

\bibitem[Ros et~al.(2016{\natexlab{a}})Ros, Sellart, Materzynska, Vazquez, and
  Lopez]{ros_cvpr16}
German Ros, Laura Sellart, Joanna Materzynska, David Vazquez, and Antonio
  Lopez.
\newblock {The SYNTHIA Dataset}: A large collection of synthetic images for
  semantic segmentation of urban scenes.
\newblock In \emph{Proc. of the IEEE Conference on Computer Vision and Pattern
  Recognition (CVPR)}, 2016{\natexlab{a}}.

\bibitem[Ros et~al.(2016{\natexlab{b}})Ros, Stent, Alcantarilla, and
  Watanabe]{ros_arxiv16}
Germ{\'{a}}n Ros, Simon Stent, Pablo~F. Alcantarilla, and Tomoki Watanabe.
\newblock Training constrained deconvolutional networks for road scene semantic
  segmentation.
\newblock \emph{CoRR}, abs/1604.01545, 2016{\natexlab{b}}.
\newblock URL \url{http://arxiv.org/abs/1604.01545}.

\bibitem[Saenko et~al.(2010)Saenko, Kulis, Fritz, and Darrell]{saenko_eccv10}
Kate Saenko, Brian Kulis, Mario Fritz, and Trevor Darrell.
\newblock Adapting visual category models to new domains.
\newblock In \emph{European conference on computer vision}, pp.\  213--226.
  Springer, 2010.

\bibitem[Salimans et~al.(2016{\natexlab{a}})Salimans, Goodfellow, Zaremba,
  Cheung, Radford, and Chen]{salimans2016improved}
Tim Salimans, Ian Goodfellow, Wojciech Zaremba, Vicki Cheung, Alec Radford, and
  Xi~Chen.
\newblock Improved techniques for training gans.
\newblock \emph{arXiv preprint arXiv:1606.03498}, 2016{\natexlab{a}}.

\bibitem[Salimans et~al.(2016{\natexlab{b}})Salimans, Goodfellow, Zaremba,
  Cheung, Radford, and Chen]{salimans_arxiv16}
Tim Salimans, Ian~J. Goodfellow, Wojciech Zaremba, Vicki Cheung, Alec Radford,
  and Xi~Chen.
\newblock Improved techniques for training gans.
\newblock \emph{CoRR}, abs/1606.03498, 2016{\natexlab{b}}.
\newblock URL \url{http://arxiv.org/abs/1606.03498}.

\bibitem[Sangkloy et~al.(2016)Sangkloy, Lu, Fang, Yu, and
  Hays]{sangkloy2016scribbler}
Patsorn Sangkloy, Jingwan Lu, Chen Fang, Fisher Yu, and James Hays.
\newblock Scribbler: Controlling deep image synthesis with sketch and color.
\newblock \emph{arXiv preprint arXiv:1612.00835}, 2016.

\bibitem[Shrivastava et~al.(2017)Shrivastava, Pfister, Tuzel, Susskind, Wang,
  and Webb]{shrivastava_cvpr17}
Ashish Shrivastava, Tomas Pfister, Oncel Tuzel, Joshua Susskind, Wenda Wang,
  and Russell Webb.
\newblock Learning from simulated and unsupervised images through adversarial
  training.
\newblock In \emph{The IEEE Conference on Computer Vision and Pattern
  Recognition (CVPR)}, July 2017.

\bibitem[Sun \& Saenko(2016)Sun and Saenko]{sun_taskcv16}
Baochen Sun and Kate Saenko.
\newblock Deep {CORAL:} correlation alignment for deep domain adaptation.
\newblock In \emph{ICCV workshop on Transferring and Adapting Source Knowledge
  in Computer Vision (TASK-CV)}, 2016.

\bibitem[Taigman et~al.(2017{\natexlab{a}})Taigman, Polyak, and Wolf]{dtn}
Y.~Taigman, A.~Polyak, and L.~Wolf.
\newblock Unsupervised cross-domain image generation.
\newblock In \emph{International Conference on Learning Representations},
  2017{\natexlab{a}}.

\bibitem[Taigman et~al.(2017{\natexlab{b}})Taigman, Polyak, and
  Wolf]{taigman_iclr17}
Yaniv Taigman, Adam Polyak, and Lior Wolf.
\newblock Unsupervised cross-domain image generation.
\newblock In \emph{International Conference on Learning Representations
  ({ICLR})}, 2017{\natexlab{b}}.

\bibitem[Torralba \& Efros(2011)Torralba and Efros]{efros_cvpr11}
Antonio Torralba and Alexei~A. Efros.
\newblock Unbiased look at dataset bias.
\newblock In \emph{CVPR'11}, June 2011.

\bibitem[Tzeng et~al.(2014)Tzeng, Hoffman, Zhang, Saenko, and
  Darrell]{tzeng_arxiv15}
Eric Tzeng, Judy Hoffman, Ning Zhang, Kate Saenko, and Trevor Darrell.
\newblock Deep domain confusion: Maximizing for domain invariance.
\newblock \emph{CoRR}, abs/1412.3474, 2014.
\newblock URL \url{http://arxiv.org/abs/1412.3474}.

\bibitem[Tzeng et~al.(2015)Tzeng, Hoffman, Darrell, and Saenko]{tzeng_iccv15}
Eric Tzeng, Judy Hoffman, Trevor Darrell, and Kate Saenko.
\newblock Simultaneous deep transfer across domains and tasks.
\newblock In \emph{International Conference in Computer Vision (ICCV)}, 2015.

\bibitem[Tzeng et~al.(2017)Tzeng, Hoffman, Saenko, and Darrell]{tzeng_cvpr17}
Eric Tzeng, Judy Hoffman, Kate Saenko, and Trevor Darrell.
\newblock Adversarial discriminative domain adaptation.
\newblock In \emph{Computer Vision and Pattern Recognition (CVPR)}, 2017.
\newblock URL \url{http://arxiv.org/abs/1702.05464}.

\bibitem[Yi et~al.(2017)Yi, Zhang, Gong, et~al.]{yi2017dualgan}
Zili Yi, Hao Zhang, Ping~Tan Gong, et~al.
\newblock Dualgan: Unsupervised dual learning for image-to-image translation.
\newblock \emph{arXiv preprint arXiv:1704.02510}, 2017.

\bibitem[Yoo et~al.(2016)Yoo, Kim, Park, Paek, and Kweon]{yoo_eccv16}
Donggeun Yoo, Namil Kim, Sunggyun Park, Anthony~S. Paek, and In{-}So Kweon.
\newblock Pixel-level domain transfer.
\newblock In \emph{European Conference on Computer Vision (ECCV)}, 2016.
\newblock URL \url{http://arxiv.org/abs/1603.07442}.

\bibitem[Yu et~al.(2017)Yu, Koltun, and Funkhouser]{drn}
Fisher Yu, Vladlen Koltun, and Thomas Funkhouser.
\newblock Dilated residual networks.
\newblock In \emph{Computer Vision and Pattern Recognition ({CVPR})}, 2017.

\bibitem[Zhang et~al.(2017)Zhang, David, and Gong]{zhang_iccv17}
Yang Zhang, Philip David, and Boqing Gong.
\newblock Curriculum domain adaptation for semantic segmentation of urban
  scenes.
\newblock In \emph{The IEEE International Conference on Computer Vision
  (ICCV)}, Oct 2017.

\bibitem[Zhao et~al.(2016)Zhao, Mathieu, and LeCun]{zhao2016energy}
Junbo Zhao, Michael Mathieu, and Yann LeCun.
\newblock Energy-based generative adversarial network.
\newblock \emph{arXiv preprint arXiv:1609.03126}, 2016.

\bibitem[Zhu et~al.(2016)Zhu, Kr{\"a}henb{\"u}hl, Shechtman, and
  Efros]{zhu2016generative}
Jun-Yan Zhu, Philipp Kr{\"a}henb{\"u}hl, Eli Shechtman, and Alexei~A. Efros.
\newblock Generative visual manipulation on the natural image manifold.
\newblock In \emph{European Conference on Computer Vision (ECCV)}, 2016.

\bibitem[Zhu et~al.(2017)Zhu, Park, Isola, and Efros]{zhu_arxiv17}
Jun-Yan Zhu, Taesung Park, Phillip Isola, and Alexei~A Efros.
\newblock Unpaired image-to-image translation using cycle-consistent
  adversarial networks.
\newblock In \emph{International Conference on Computer Vision ({ICCV})}, 2017.

\end{thebibliography}
